\documentclass[11pt]{article}

\usepackage[final]{acl}

\usepackage{times}
\usepackage{latexsym}
\usepackage{svg}
\usepackage{tabularray}
\usepackage{multirow}
\usepackage{subfig}
\usepackage[T1]{fontenc}

\usepackage[utf8]{inputenc}

\usepackage{microtype}

\usepackage{inconsolata}

\usepackage{graphicx}

\title{Segmentation Beyond Defaults: Asymmetrical Byte Pair Encoding for Optimal Machine Translation Performance}
\author{Saumitra Yadav \and Manish Shrivastava \\
        Language Technologies Research Center, KCIS,\\ International Institute of Information Technology Hyderabad, India
        \\
        \texttt{saumitra.yadav@research.iiit.ac.in} \and \texttt{m.shrivastava@iiit.ac.in} 
        }

\begin{document}
\maketitle
\begin{abstract}
Existing Machine Translation (MT) research often suggests a single, fixed set of hyperparameters for word segmentation models, \textbf{symmetric Byte Pair Encoding} (BPE), which applies the same number of merge operations (NMO) to train tokenizers for both source and target languages. However, we demonstrate that this uniform approach doesn't guarantee optimal MT performance across different language pairs and data sizes. This work investigates BPE segmentation recipes across various data volumes and language pairs to evaluate MT system performance. We find that utilizing \textbf{asymmetric BPE}—where the source and target languages have different NMOs—significantly improves results over the symmetric approach, especially in low-resource settings (50K, 100K, and 500K sentence pairs). Specifically, asymmetric BPE yield statistically significant (p<0.05) average gains of 5.32, 4.46, and 0.7 CHRF++ on English-Hindi in low-resource setups (50K, 100K, and 500K sentence pairs, respectively). We validated this trend across six additional language pairs (English$\leftrightarrow$Telugu, Shona, Norwegian, Kyrgyz, Hausa, and Inuktitut), observing statistically significant improvement in 10 out of 12 systems compared to symmetric BPE. Our findings indicate a high NMO for the source (4K to 32K) and a low NMO for the target (0.5K to 2K) provides optimal results, particularly benefiting low-resource MT.
\end{abstract}

\section{Introduction}
Efforts have been made to include low-resource language pairs in Neural Machine Translation (NMT), e.g. \href{https://aclanthology.org/venues/loresmt/}{Workshop on Technologies for MT of Low Resource Languages}. Often, successful past methodologies on high-resource language pairs, like hyperparameters for preprocessing, are used without considering their suitability for specific language pairs. For example, if we take a preprocessing step, such as word segmentation, a key preprocessing step, divides words into subwords to enhance learning and manage vocabulary size, handling rare and unknown words to boost MT performance. Notable Techniques include BPE \cite{sennrich-etal-2016-neural}, word piece \cite{devlin2019bertpretrainingdeepbidirectional}, sentence piece \cite{kudo-richardson-2018-sentencepiece}, and morfessor \cite{smit-etal-2014-morfessor}. BPE compresses data by merging frequent character pairs into symbols \cite{gage1994new}, with the \textit{number of merge operations} (NMO) as a key parameter. 
A lower NMO (e.g., 500, Table~\ref{tab:bpe1}) reduces vocabulary size with more segmentation, while a higher NMO (e.g., 32K) results in larger vocabularies and less segmentation. Typically, the same NMO is applied to both source and target languages. Recent work have shown that examining BPE parameters in low-resource MT is vital \cite{ding-etal-2019-call,abid-2020-sadid}, but uniform NMOs for source and target (symmetrical BPE) \cite{huck-etal-2017-target,ortega2020neural,lankford-etal-2021-transformers,10.1007/978-3-031-24337-0_38,lee-etal-2024-length} prevail, with little exploration of asymmetrical BPE in MT. Earlier work \citet{ngo-ho-yvon-2021-optimizing} looked at asymmetric BPE for language alignment, not for MT. Our work is a result of a multi-year exploration of the impact of asymmetrical subword segmentation in bilingual MT systems.
\begin{table*}[h!]
\centering

\begin{tabular}{|l|l|}
\hline
Sentence &
  \begin{tabular}[c]{@{}l@{}}bosusco , 54 , runs an adventure tourism bureau .\end{tabular} \\ \hline
500 NMO &
  \begin{tabular}[c]{@{}l@{}}bo@@ su@@ sc@@ o , 5@@ 4 , r@@ un@@ s an \\ ad@@ v@@ en@@ ture t@@ our@@ is@@ m bu@@ re@@ a@@ u .\end{tabular} \\ \hline
32K NMO &
  \begin{tabular}[c]{@{}l@{}}bo@@ su@@ sco , 54 , runs an adventure tourism bureau .\end{tabular} \\ \hline
\end{tabular}
\caption{Effect of NMO variation: 500 NMO yields highly segmented tokens, while 32K retains most vocabulary}\label{tab:bpe1}
\end{table*}

While we acknowledge the rise of multilingual and decoder-only models, our study focuses on the effect of asymmetric BPE in bilingual setups, particularly in low-resource conditions where pretrained tokenizers or joint vocabularies may be unavailable. Bilingual systems remain a research focus, with studies in Cantonese-Mandarin \cite{liu-2022-low}, English-Luganda \cite{10.1145/3711542.3711594}, Wolof-French \cite{dione-etal-2022-low}, Bavarian-German \cite{her-kruschwitz-2024-investigating}, and English-Manipuri \cite{singh-etal-2023-nits,10.1016/j.eswa.2022.118187} using bilingual data and transformer-based architectures with customized subword segmentation like BPE or morphology-aware tokenization. These efforts, along with \citet{10.1007/978-981-97-5669-8_4}, cover underrepresented languages and diverse writing systems, proving the continued relevance of bilingual systems. Our work investigates asymmetrical BPE's impact on bilingual MT systems, utilizing different merge operation counts for source and target languages across varied dataset sizes and resources. Extending these results to multilingual or decoder-only models is beyond this work's scope but represents an interesting future direction.

We define the ``BPE configuration" as $m_1$\_$m_2$, with $m_1$ and $m_2$ representing the merge operations for source and target languages. Our study on symmetric and asymmetric BPE configurations for English–Hindi under varying data conditions shows asymmetric configurations performing best, especially in low-resource context. We extend these insights to six additional language pairs—English $\leftrightarrow$ {Telugu, Shona, Norwegian, Kyrgyz, Hausa, Inuktitut}—selected for diverse language families and morphological typologies. \textbf{Our findings consistently demonstrate that, in low-resource environments, the most effective BPE configuration for the majority of language translation directions tends to be asymmetric. Specifically, setups with \textit{4K} to \textit{32K} NMO for the source and \textit{500} to \textit{2K} for the target outperform symmetric BPE configurations.}

Section \ref{sec:relatedWork} summarizes previous efforts to use symmetric BPE merge operations to improve MT performance. Section \ref{sec:exploring} explains our motivation for finding optimal BPE configurations by exploring asymmetric BPE. Section \ref{sec:expsetup} outlines our experimental setup and presents the performance of the English-Hindi MT system on FLORES and Domain testsets. Section \ref{sec:others} evaluates the setup for other language pairs in low resource context, concluding our observations in Section \ref{sec:conclusion}.

\section{Related Work - Symmetrical BPE}
\label{sec:relatedWork}
Most bilingual MT systems—especially for low-resource pairs—use the same number of merge operations (NMO) for source and target languages. Studies show that smaller vocabularies (0–4K NMO) outperform the common 32K setting by up to 4 BLEU points in low-resource scenarios~\cite{ding-etal-2019-call}; similar patterns are reported for English–Egyptian, English–Levantine~\cite{abid-2020-sadid}, and English–Irish~\cite{lankford-etal-2021-transformers}.

Other work adapts segmentation for polysynthetic languages~\cite{ortega2020neural}, rich morphology~\cite{lee-etal-2024-length}, or target-side variation~\cite{10.1007/978-3-031-24337-0_38}. Alternative strategies include cascading segmentations~\cite{huck-etal-2017-target}, vocabulary refinement~\cite{xu-etal-2021-vocabulary}, and multi-BPE–setting corpora~\cite{poncelas-etal-2020-using}. While~\cite{ngo-ho-yvon-2021-optimizing} varied NMOs for alignment, no prior study systematically evaluates asymmetric BPE—using different NMOs for source and target—across resource levels. This work addresses that gap.

Though multilingual MT research now dominates, bilingual MT remains vital for low-resource pairs, where symmetric BPE is still common~\cite{liu-2022-low,10.1145/3711542.3711594,dione-etal-2022-low,her-kruschwitz-2024-investigating,singh-etal-2023-nits,10.1016/j.eswa.2022.118187}. Recent work on Parity-Aware BPE \cite{foroutan2025parityawarebytepairencodingimproving} introduces fairness-oriented subword allocation, reducing disadvantages for low-resource languages in multilingual tokenization. Although our experiments are limited to bilingual MT, asymmetric BPE could complement such fairness-aware methods in multilingual systems; extending this remains outside our current scope.

\section{Exploring Asymmetrical BPE}
\label{sec:exploring}

\begin{figure}[]
 \centering
  \includegraphics[width=\columnwidth]{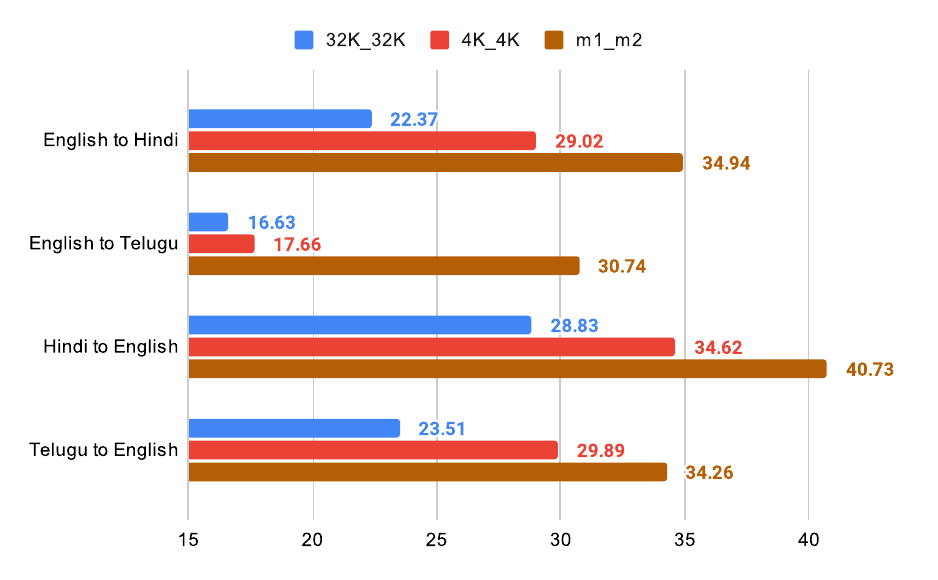}
 \caption {CHRF++ Scores for Symmetrical BPE (\textit{32K},\textit{4K}) vs Asymmetrical BPE ($m_1$ $\neq$ $m_2$)}
  
  \label{fig:atypicalBPE}
\end{figure}

In practice, for a BPE configuration $m_1$\_$m_2$, the values of $m_1$ and $m_2$ are usually the same, with the number of merge operations (NMO) ranging from \textit{8K} to \textit{40K}~\cite{wu2016google,denkowski-neubig-2017-stronger,cherry-etal-2018-revisiting,renduchintala-etal-2019-character}. However, \citet{ding-etal-2019-call,dewangan2021experience} found these settings suboptimal for low-resource language pairs. \citet{ding-etal-2019-call} observed that $m_1 = m_2 \leq \textit{4K}$ NMO outperforms \textit{32K} in low-resource conditions, consistent with our experiments on 0.1 million sentence pairs (English $\leftrightarrow$ \{Hindi, Telugu\}) (Figure~\ref{fig:atypicalBPE}). \citet{dewangan2021experience} further showed that identical BPE configurations yield differing performance across language pairs, exemplified by English-Hindi vs. English-Telugu comparisons at \textit{4K} NMO (Figure~\ref{fig:atypicalBPE}).

\begin{figure*}[!t]
 \centering
  \includegraphics[width=0.80\linewidth]{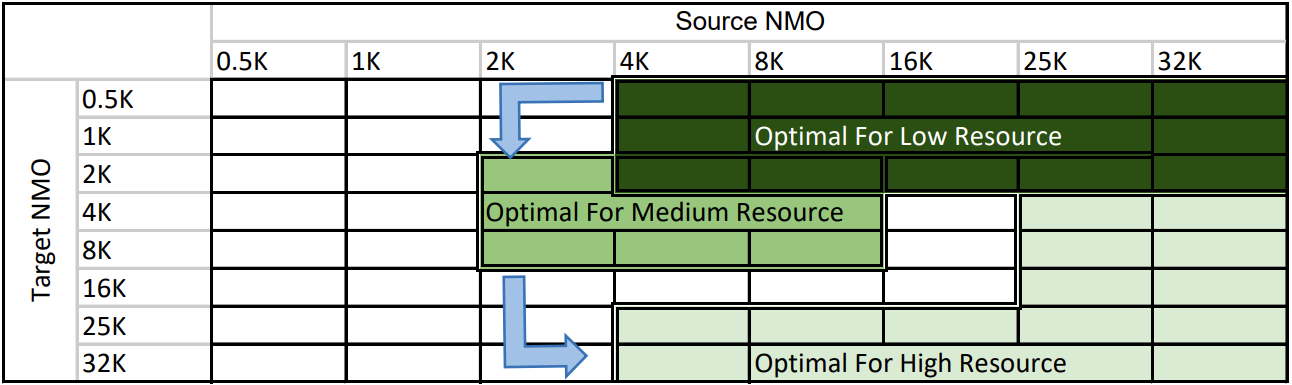}
 \caption {Changes in Optimal BPE Configuration from Low- to High-Resource Settings}
  
  \label{fig:generally}
\end{figure*}
Work by \citet{ortega2020neural,mujadia-sharma-2021-english} suggests that selecting NMO should be done while considering dataset size and language pair, as nuanced BPE strategies benefit morphologically complex languages. We study symmetrical BPE configurations with identical NMOs for source and target, and investigate alternatives by varying $m_1$ and $m_2$ independently in English-Hindi across datasets from 50K to 8M sentences. This approach improves results in low-resource settings (Figure~\ref{fig:atypicalBPE}). Extensive experiments on English-Hindi, evaluated on FLORES~\cite{goyal-etal-2022-flores}, confirm better performance of atypical BPE for tokenization. We further validate these findings by extending experiments to English $\leftrightarrow$ \{Telugu, Shona, Norwegian, Kyrgyz, Hausa, Inuktitut\}.
\begin{table*}[]
\centering
\resizebox{0.9\textwidth}{!}{%
\begin{tabular}{|c|c|c|c|c|c|c|c|c|c|}
\hline
Length bin & 1 to 10 & 11 to 15 & 16 to 20 & 21 to 25 & 26 to 30 & 31 to 35 & 35 to 40 & \textgreater{}=41 & Total \\ \hline
No. of sentences & 2792334 & 1655162 & 1150396 & 854091 & 617318 & 420583 & 275774 & 414926 & 8180584 \\ \hline
Percentage & 34.13 & 20.23 & 14.06 & 10.44 & 7.55 & 5.14 & 3.37 & 5.07 & 100 \\ \hline
\end{tabular}%
}
\caption{Distribution of sentences in groups based on token length for full data}
\label{tab:sampling}
\end{table*}
Our results strongly support optimizing NMO based on training data size and language pair. Figure~\ref{fig:generally} presents a conceptual overview of the \textbf{optimal ranges} for \textbf{BPE configurations} found in English-Hindi across resource settings. Here, ``ranges'' indicate the spectrum of NMO values used as hyperparameters for source and target subword tokenization in word segmentation. The performance gap between the best and symmetrical BPE systems is shown by shades of green, with the largest gains in low-resource scenarios (darker green). As dataset size increases, performance differences among configurations diminish (lighter green).


\section{Evaluation on English $\leftrightarrow$ Hindi}
\label{sec:expsetup}
We explore BPE configurations with the Samanantar dataset~\cite{10.1162/tacl_a_00452} for English-Hindi
containing 8 million parallel sentences. English text is tokenized, normalized, and lowercased using Moses scripts\footnote{\url{https://github.com/moses-smt/mosesdecoder/}}, while preprocessing of Hindi utilizes the Indic NLP library~\cite{kunchukuttan2020indicnlp}. We simulate various training set sizes by grouping sentences based on English sentence length (Table~\ref{tab:sampling}) and randomly sample datasets of sizes 0.05M, 0.1M, 0.5M, 1M, 4M, and 8M, maintaining sentence length proportions (see Appendix~\ref{subsec:enhi_stats} for details). The BPE tokenizer is trained per language and dataset size with eight NMOs: \textit{0.5K}, \textit{1K}, \textit{2K}, \textit{4K}, \textit{8K}, \textit{16K}, \textit{25K}, and \textit{32K}. 

All possible BPE configurations (e.g., src$_{500}$-tgt$_{500}$, src$_{500}$-tgt$_{1000}$) are trained using the Transformer architecture~\cite{vaswani2017attention} with hyperparameters detailed in Appendix~\ref{subsec:hyperparameters}. Training a single BPE configuration $m_1$\_$m_2$ across all dataset sizes averages 1040 GPU hours on a 1080TI, resulting in 64 configurations per language direction and 768 total systems (64 configurations $\times$ 6 dataset sizes $\times$ 2 directions). For evaluation, we use the FLORES dataset~\cite{goyal-etal-2022-flores} and report CHRF++ scores~\cite{popovic-2015-chrf} to analyze the impact of different BPE configurations. We adopt CHRF++ rather than embedding-based metrics such as COMET \cite{rei-etal-2022-cometkiwi}, as not all language pairs have COMET support and we aim to compare performance using a consistent metric across all pairs. Validation and test set statistics are provided in Appendix~\ref{subsec:validAndTest}.

\subsection{Best and Worst Configurations}
To maintain clarity and brevity in our observations, Tables \ref{tab:hin2eng} and Table \ref{tab:eng2hin_neat} show the performance of five selected configurations out of 64. For each dataset size, the systems represented are:
\begin{itemize}
    \item High A and B: The two systems with the highest performance across all asymmetric configurations for each dataset size.
    \item Low A and B: The two systems with the lowest performance across all asymmetric configurations for each dataset size.
    \item Baseline: The best system among all symmetric BPE configurations (\textit{m\_m}, where \textit{m}$\subset$\{\textit{500},\textit{1K},\textit{2K},\textit{4K},\textit{8K},\textit{16K},\textit{25K},\textit{32K}\}).
\end{itemize}
Performance of all configurations for all systems is provided in the Appendix \ref{subsec:allSystemsENHI}.
\begin{table*}
\centering
\resizebox{\textwidth}{!}{%
\begin{tblr}{
  cells = {c},
  cell{1}{2} = {c=4}{},
  cell{1}{6} = {c=4}{},
  cell{1}{10} = {c=4}{},
  cell{8}{2} = {c=4}{},
  cell{8}{6} = {c=4}{},
  cell{8}{10} = {c=4}{},
  vline{2-3,7} = {1,8}{},
  vline{2,6,10} = {2-7,9-14}{},
  hline{1-3,8-10,15} = {-}{},
}
Dataset Size           & 0.05 M &     &                 &       & 0.1 M &     &                 &        & 0.5 M &     &                 &       \\
Performance Tier & src    & tgt & CHRF++          & $\delta$ & src   & tgt & CHRF++          & $\delta$  & src   & tgt & CHRF++          & $\delta$ \\
Low A                  & 500     & 1K & 19.56           & -3.93 & 500   & 25K & 23.36           & -15.92 & 2K    & 32K & 48.92           & -3.53 \\
Low B                  & 500     & 2K & 19.58           & -3.91  & 1K    & 32K & 24.2            & -15.08 & 25K   & 32K & 49.62           & -2.83 \\
Baseline               & 4K     & 4K  & 23.49           & 0     & 500   & 500 & 39.28           & 0      & 4K    & 4K  & 52.45           & 0     \\
High B                 & 25K    & 500 & \textbf{28.47*} & 4.98  & 16K   & 500 & \textbf{40.66*} & 1.38   & 8K    & 2K  & \textbf{53.19*} & 0.74  \\
High A                 & 16K    & 500 & \textbf{29.33*} & 5.84  & 8K    & 500 & \textbf{40.75*} & 1.47   & 4K    & 500 & \textbf{53.37*} & 0.92  \\
Dataset Size           & 1 M    &     &                 &       & 4 M   &     &                 &        & 8 M   &     &                 &       \\
Performance Tier & src    & tgt & CHRF++          & $\delta$ & src   & tgt & CHRF++          & $\delta$  & src   & tgt & CHRF++          & $\delta$ \\
Low A                  & 500    & 32K & 53.27           & -1.77 & 500   & 1K  & 56.1            & -1.73  & 500   & 2K  & 56.26           & -2.45 \\
Low B                  & 1K     & 32K & 53.58           & -1.46 & 1K    & 2K  & 56.3            & -1.53  & 500   & 500 & 56.43           & -2.28 \\
Baseline               & 8K     & 8K  & 55.04           & 0     & 32K   & 32K & 57.83           & 0      & 32K   & 32K & 58.71           & 0     \\
High B                 & 16K    & 8K  & 55.19           & 0.15  & 32K   & 16K & 58.06           & 0.23   & 16K   & 25K & 58.74           & 0.03  \\
High A                 & 16K    & 4K  & 55.39           & 0.35  & 25K   & 16K & 58.18           & 0.35   & 4K    & 32K & 58.75           & 0.04  
\end{tblr}}

\caption{Performance of the top 2 (High A, High B) and bottom 2 (Low A, Low B) tokenization configurations compared to the symmetric baseline for Hindi-to-English across dataset sizes. Bold indicates statistically significant improvement over baseline ($p < 0.05$); bold with * denotes high significance ($p < 0.01$). $\delta$ shows CHRF++ difference from best baseline. \textbf{src} and \textbf{tgt} are source and target merge operations (NMO).}
\label{tab:hin2eng}
\end{table*}

\begin{figure}
\includegraphics[width=0.48\textwidth]{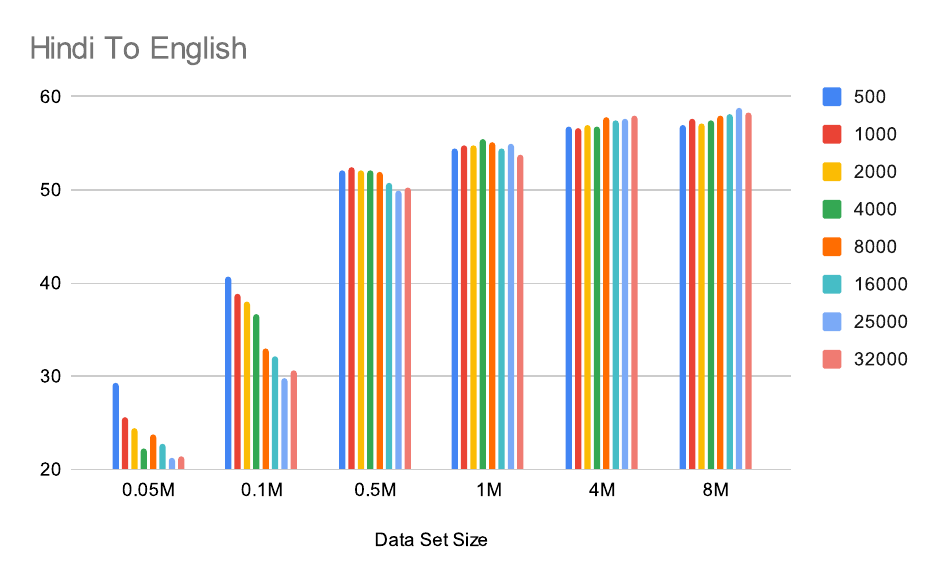}
\caption{CHRF++ scores for 0.1M sentence pairs for \textit{Hindi-to-English} MT systems using configurations of the form \textit{16K}\_\textit{x}, where $x \in$ \{\textit{500}, \textit{1K}, \textit{2K}, \textit{4K}, \textit{8K}, \textit{16K}, \textit{25K}, \textit{32K}\}.}
\label{fig:only16KBPEhin2eng}
\end{figure}

\begin{figure}
\centering
\includegraphics[width=0.48\textwidth]{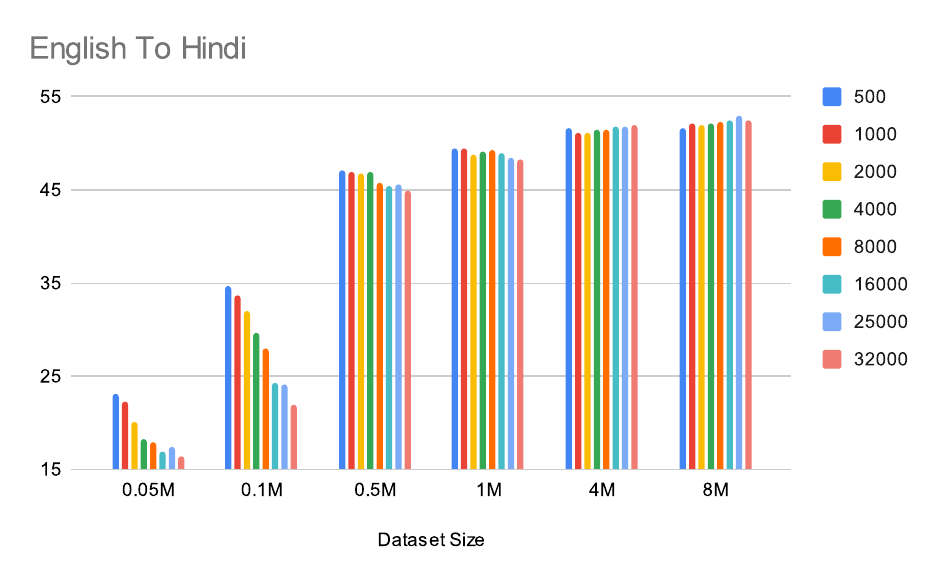}
\caption{CHRF++ scores for 0.1M sentence pairs for \textit{English-to-Hindi} MT systems using configurations of the form \textit{16K}\_\textit{x}, where $x \in$ \{\textit{500}, \textit{1K}, \textit{2K}, \textit{4K}, \textit{8K}, \textit{16K}, \textit{25K}, \textit{32K}\}.}
\label{fig:only16KBPEeng2hin}
\end{figure}

\begin{table*}
\centering
\resizebox{\textwidth}{!}{%
\begin{tblr}{
  cells = {c},
  cell{1}{2} = {c=4}{},
  cell{1}{6} = {c=4}{},
  cell{1}{10} = {c=4}{},
  cell{8}{2} = {c=4}{},
  cell{8}{6} = {c=4}{},
  cell{8}{10} = {c=4}{},
  vline{2-3,7} = {1,8}{},
  vline{2,6,10} = {2-7,9-14}{},
  hline{1-3,8-10,15} = {-}{},
}
Dataset Size     & 0.05 M &     &                 &       & 0.1 M &      &                 &        & 0.5 M &     &                &       \\
Performance Tier & src    & tgt & CHRF++          & $\delta$ & src   & tgt  & CHRF++          & $\delta$  & src   & tgt & CHRF++         & $\delta$ \\
Low A            & 1K     & 25K & 13           & -5.39 & 500   & 32K   & 16.49           & -12.55  & 500   & 32K & 43.57          & -3.5  \\
Low B            & 500     & 4K & 13.55              & -4.84 & 500   & 25K  & 16.74           & -12.3 & 1K    & 32K & 43.88          & -3.19 \\
Baseline         & 8K     & 8K  & 18.39           & 0     & 4K    & 4K   & 29.04           & 0      & 4K    & 4K  & 47.07          & 0     \\
High B           & 16K    & 500 & \textbf{23.19*} & 4.8   & 16K   & 500  & \textbf{34.73*} & 5.69   & 8K    & 500 & 47.12          & 0.05  \\
High A           & 8K     & 500 & \textbf{23.83*} & 5.44  & 8K    & 500  & \textbf{35*}    & 5.96   & 4K    & 500 & \textbf{47.55} & 0.48  \\
Dataset Size     & 1 M    &     &                 &       & 4 M   &      &                 &        & 8 M   &     &                &       \\
Performance Tier & src    & tgt & CHRF++          & $\delta$ & src   & tgt  & CHRF++          & $\delta$  & src   & tgt & CHRF++         & $\delta$ \\
Low A            & 1K     & 32K & 47.23           & -1.93 & 8K    & 2K   & 50.64           & -1.12  & 500   & 1K  & 50.79          & -1.84 \\
Low B            & 2K     & 32K & 47.83           & -1.33 & 500  & 2K & 50.73           & -1.03  & 32K   & 2K & 51.29           & -1.34 \\
Baseline         & 8K     & 8K  & 49.16           & 0     & 16K   & 16K  & 51.76           & 0      & 25K   & 25K & 52.63          & 0     \\
High B           & 4K     & 2K  & \textbf{49.74}  & 0.58  & 16K   & 32K  & 51.95           & 0.19   & 25K   & 32K & 52.63          & 0     \\
High A           & 8K     & 2K  & \textbf{49.75}  & 0.59  & 32K   & 25K  & 52              & 0.24   & 16K   & 25K & \textbf{53}    & 0.37  
\end{tblr}}
\caption{Performance of the top 2 (High A, High B) and bottom 2 (Low A, Low B) tokenization configurations compared to the symmetric baseline for English-to-Hindi across dataset sizes. Bold indicates statistically significant improvement over baseline ($p < 0.05$); bold with * denotes high significance ($p < 0.01$). $\delta$ shows CHRF++ difference from best baseline. \textbf{src} and \textbf{tgt} are source and target merge operations (NMO).}
\label{tab:eng2hin_neat}
\end{table*}
As shown in Tables~\ref{tab:hin2eng} and~\ref{tab:eng2hin_neat}, for low-resource settings ($<$1M), the best system outperforms the weakest by $\approx$15 CHRF++ scores and the best symmetric BPE by $\approx$5. In medium-resource scenarios (1M), the optimal source and target NMO shift to the medium range (2K--8K), with smaller performance variation ($\approx$3 CHRF++). For high-resource settings, the difference between best and worst configurations is minimal (< 2 CHRF++), with the best system using 32K NMO on the target. This highlights the advantage of asymmetric BPE in low-resource contexts. This trend of shifting optimal BPE values with dataset size also appears when varying target NMO while keeping source NMO fixed. For example, English$\leftrightarrow$Hindi systems with source NMO fixed at 16K on 0.1M data (Figures~\ref{fig:only16KBPEhin2eng} and~\ref{fig:only16KBPEeng2hin}) show gradual performance changes as target NMO varies from 500 to 32K. Similar patterns with other fixed source or target values are detailed in Appendix~\ref{subsec:allSystemsENHI}. This highlights that modifying the NMO on the target side, especially in a low-resource scenario, plays a vital role in determining the optimal BPE configuration.

We conclusively find that symmetric BPE configurations underperform compared to asymmetric ones in low-resource MT systems. As dataset size grows, symmetric configurations perform comparably to asymmetric. Nonetheless, asymmetric BPE yields statistically significant improvements in low-resource settings.
\begin{figure}
\centering
\includegraphics[width=0.48\textwidth]{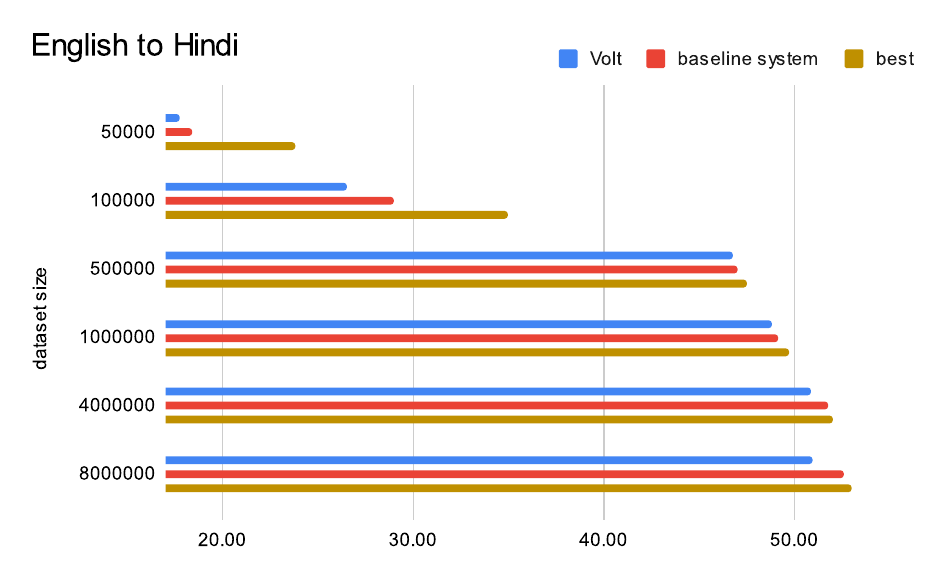}
\caption{CHRF++ score comparison of Asymmetric BPE with VOLT for English to Hindi}\label{fig:volt_en2hi}
\end{figure}

\begin{figure}
\centering
\includegraphics[width=0.48\textwidth]{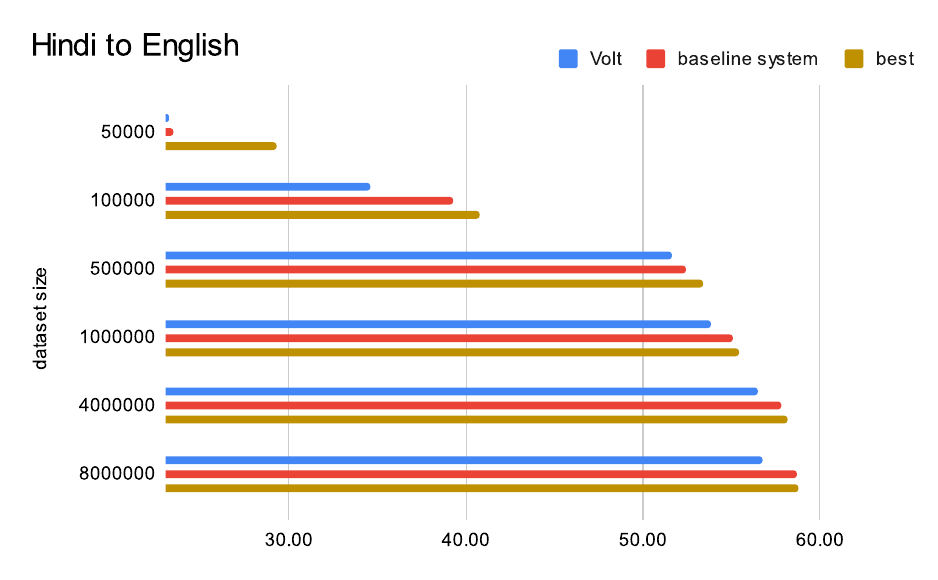}
\caption{CHRF++ score comparison of Asymmetric BPE with VOLT for Hindi to English}\label{fig:volt_hi2en}
\end{figure}

We compare our systems with optimal BPE configurations against VOLT~\cite{xu-etal-2021-vocabulary}\footnote{Using hyperparameters specified in the original paper.}. Figures~\ref{fig:volt_en2hi} and~\ref{fig:volt_hi2en} show CHRF++ comparisons between VOLT tokenization, optimal BPE, and ``best" baseline symmetric BPE (source NMO = target NMO) configuration. Systems using asymmetric BPE outperform VOLT across all dataset sizes, with statistically significant improvements ($p < 0.05$) especially in low-resource settings.

\subsection{Performance on Domain Test}

Subword models must handle rare or unseen words, making domain-specific datasets effective for evaluating asymmetric BPE in MT systems.
Thus, to demonstrate the impact of segmentation strategies, we evaluate all systems on Artificial Intelligence (AI) and Chemistry (CH) domain test sets from the \href{https://www.iitp.ac.in/~ai-nlp-ml/icon2020/shared_tasks.html}{ICON 2020 Domain Adaptation Task}\footnote{We thank task organizers for access.}. Table~\ref{tab:domainStats}\footnote{After removing 12 and 5 lines from AI and CH test sets respectively, that overlapped with the 8M training set.} presents domain test data statistics. Table ~\ref{tab:domain_en2hi} show the performance of configurations from Table~\ref{tab:eng2hin_neat}  on domain datasets for English-to-Hindi systems. Performance of Hindi-to-English systems is given in Appendix~\ref {subsec:hin2eng_domain}.
\begin{table}[]
\centering
\resizebox{\columnwidth}{!}{%
\begin{tabular}{|c|c|c|c|}
\hline
\textbf{Domain} & \textbf{\# of Sentences} &\textbf{English Tokens} & \textbf{Hindi Tokens} \\ \hline
Artificial Intelligence     & 389     & 6965 & 8441 \\ \hline
Chemistry      & 392     & 7761 & 9368 \\ \hline
\end{tabular}
}
\caption{Statistics of ICON 2020 Domain Adaptation Testset}
\label{tab:domainStats}
\end{table}

For English$\leftrightarrow$Hindi domain test set translation, we observe:

\begin{itemize}
    \item In low- to medium-resource settings, asymmetric BPE systems outperform baselines significantly when source NMO is much higher than target NMO. This aligns with FLORES results (Tables~\ref{tab:hin2eng} and~\ref{tab:eng2hin_neat}) and highlights asymmetric BPE benefits for domain translation with limited data.
    \item In high-resource settings, symmetric and asymmetric systems perform similarly.
\end{itemize}

These results demonstrate the potential translation improvements from asymmetric BPE in new domains under limited-resource conditions. Performances of all systems on AI and CH test sets is in Appendices~\ref{ch:appendixAI} and~\ref{ch:appendixCH}, respectively.

\begin{table*}
\centering
\resizebox{\textwidth}{!}{%
\begin{tblr}{
  cells = {c},
  cell{1}{2} = {c=4}{},
  cell{1}{6} = {c=4}{},
  cell{1}{10} = {c=4}{},
  cell{8}{2} = {c=4}{},
  cell{8}{6} = {c=4}{},
  cell{8}{10} = {c=4}{},
  vline{2-3,7} = {1,8}{},
  vline{2,6,10} = {2-7,9-14}{},
  hline{1-3,8-10,15} = {-}{},
}
Dataset
  Size   & 0.05M &     &                 &                 & 0.1M &     &                 &                 & 0.5M &     &       &                \\
Performance Tier & src   & tgt & AI              & CH              & src  & tgt & AI              & CH              & src  & tgt & AI    & CH             \\
Low A            & 1K    & 25K & 15.98           & 14.13            & 500  & 32K  & 18.46           & 16.67           & 500  & 32K & 53.32 & 47.44          \\
Low B            & 500    & 4K & 15.97           & 15.03           & 500  & 25K & 18.80           & 16.86           & 1K   & 32K & 53.99 & 47             \\
Baseline         & 8K    & 8K  & 20.76           & 19.34           & 4K   & 4K  & 35.79           & 32.19           & 4K   & 4K  & 58.63 & 50.64          \\
High B           & 16K   & 500 & \textbf{26.76*} & \textbf{24.03*} & 16K  & 500 & \textbf{42.97*} & \textbf{37.94*} & 8K   & 500 & 58.91 & 50.94          \\
High A           & 8K    & 500 & \textbf{28.28*} & \textbf{25.14*} & 8K   & 500 & \textbf{44.05*} & \textbf{38.57*} & 4K   & 500 & 58.70 & \textbf{51.53} \\
Dataset Size     & 1M    &     &                 &                 & 4M   &     &                 &                 & 8M   &     &       &                \\
Performance Tier & src   & tgt & AI              & CH              & src  & tgt & AI              & CH              & src  & tgt & AI    & CH             \\
Low A            & 1K    & 32K & 58.58           & 51.78           & 8K   & 2K  & 62.23           & 54.55           & 500  & 1K  & 61.91 & 54.78          \\
Low B            & 2K    & 32K & 58.88           & 51.65           & 500  & 2K & 61.51           & 54.01           & 32K  & 2K & 62.52 & 54.63          \\
Baseline         & 8K    & 8K  & 61.22           & 53.6            & 16K  & 16K & 63.12           & 55.14           & 25K  & 25K & 63.95 & 55.65          \\
High B           & 4K    & 2K  & 60.39           & 53.55           & 16K  & 32K & 63.21           & \textbf{55.84}  & 25K  & 32K & 63.9  & 55.92          \\
High A           & 8K    & 2K  & 60.01           & 53.27           & 32K  & 25K & 63.6            & 55.74           & 16K  & 25K & 63.53 & 55.69          
\end{tblr}}
\caption{Performance of the top 2 (High A and High B) and bottom 2 (Low A and Low B) systems with respective tokenisation configurations compared to the symmetric baseline for \textit{English-to-Hindi} systems across dataset sizes for \textbf{AI} and \textbf{CH Domains}. Bold scores indicate statistically significant improvements over the baseline ($p\textless{}0.05$); bold scores with an asterisk ($*$) indicate high significance ($p\textless{}0.01$)}
\label{tab:domain_en2hi}
\end{table*}
\begin{figure*}
\centering
\includegraphics[width=\textwidth]{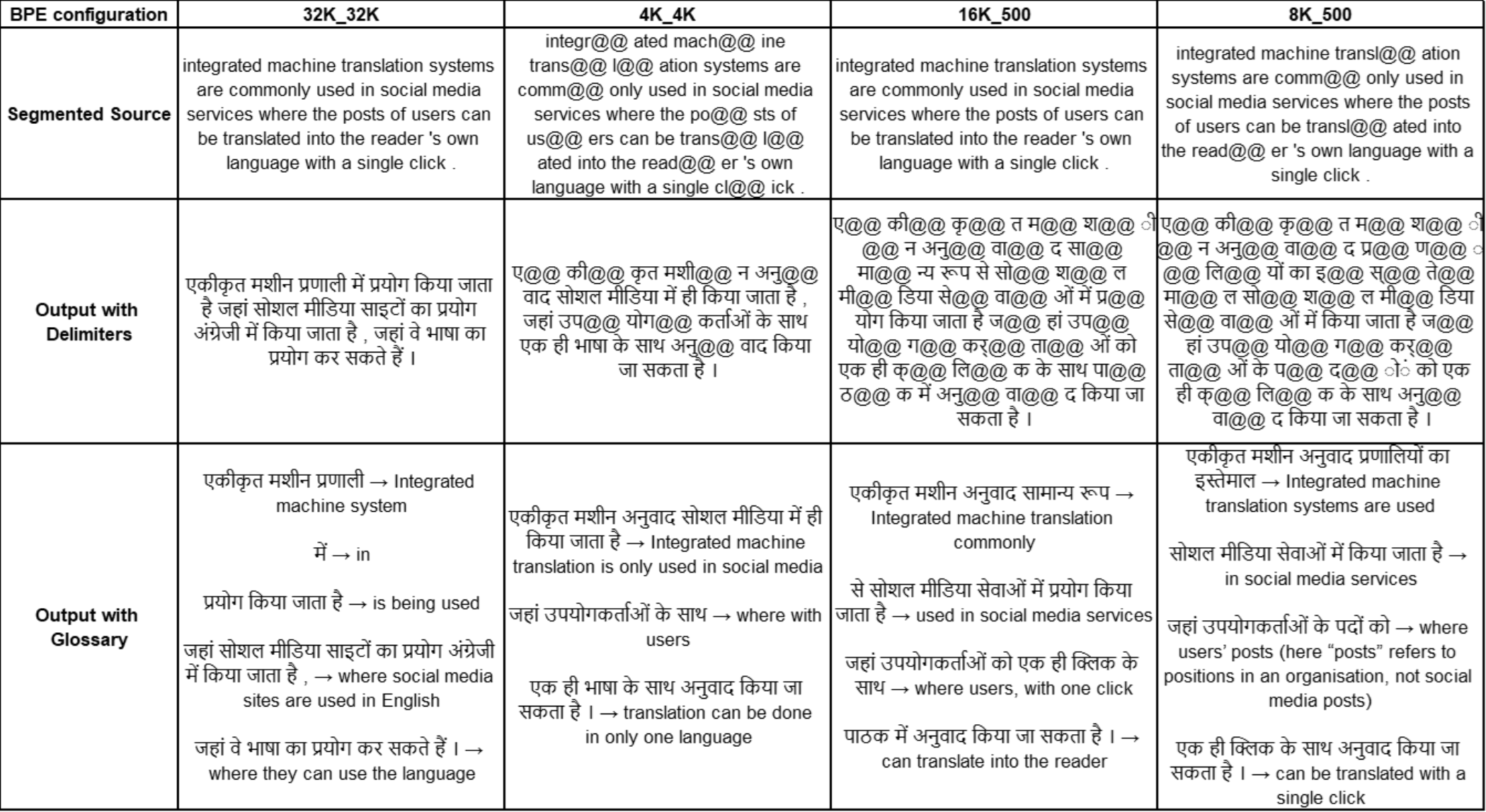}
\caption{Examples of English-to-Hindi translations across different BPE configurations, showing segmented source text, outputs with delimiters `@@', and output without delimiters with corresponding English glossaries for each segment.}\label{fig:aiExample}

\end{figure*}
Figure~\ref{fig:aiExample} illustrates, with an example on AI domain, the advantage of asymmetric BPE over symmetric BPE for 0.1M parallel sentences. Configurations like \textit{16K\_500} or \textit{8K\_500} produce more natural, semantically faithful Hindi translations than symmetric \textit{32K\_32K} or \textit{4K\_4K} setups. Translation improves as we move from symmetric high NMO (\textit{32K\_32K}), to symmetric low NMO (\textit{4K\_4K}), to asymmetric (\textit{16K\_500} or \textit{8K\_500}).

\begin{itemize}
    
\item \textbf{\textit{32K}\_\textit{32K}} – In the output with delimiters, most of the tokens are already fully merged into complete words. While this segmentation yields a large vocabulary, in low-resource conditions, it results in sparsity: many source and target tokens appear too infrequently for effective parameter learning. Consequently, the network fails to learn robust mappings, leading to incomplete or inaccurate translations despite having fully merged tokens.

\item \textbf{\textit{4K}\_\textit{4K}} – The glossary shows an improvement in overall translation fluency, but important content words such as system, commonly and click are missing, both explicitly and implicitly (meaning that they cannot be inferred from context). The improvement is due to the increased recurrence of subword units in the training data from the reduced vocabulary size, which strengthens learned associations, but at the cost of certain semantic details.
    \item \textbf{Asymmetric (\textit{16K\_500}, \textit{8K\_500})}: Better meaning preservation than symmetric. Whereas \textit{16K\_500} omits “post” and drops final language reference, \textit{8K\_500} conveys almost full meaning but mistranslates ``post'' as a job title. From a learning perspective, the smaller decoder vocabulary improves the alignment and connection learning between the source and target segments (similar to \citet{ngo-ho-yvon-2021-optimizing}), aligning with previous findings \cite{10.1007/978-3-031-24337-0_38} that the target side vocabulary influences NMT performance. Although overly constrained vocabularies can still introduce semantic drift in rare or domain-specific terms, overall translation remains improved compared to symmetric configurations.
\end{itemize}

\section{Exploring Asymmetrical BPE Configurations for other language pairs}
\label{sec:others}

To evaluate the transferability of optimal subword segmentation from English–Hindi to typologically diverse languages, we extend experiments to English$\leftrightarrow$\{Telugu, Shona, Norwegian, Kyrgyz, Hausa, Inuktitut\}. Corpora sources are:
\begin{itemize}
    \item \textbf{English–\{Hausa, Shona, Norwegian, Kyrgyz\}}: \citet{gowda-etal-2021-many}
    \item \textbf{English–Telugu}: \citet{10.1162/tacl_a_00452}
    \item \textbf{English–Inuktitut}: \citet{joanis-etal-2020-nunavut}
\end{itemize}

To simulate low-resource settings, we sampled 0.1M sentence pairs per language via sentence-length binning, analogous to English–Hindi, statistics are in Appendix~\ref{subsec:stats_others}. 

These language pairs were chosen to assess the impact of symmetric and asymmetric BPE configurations in low-resource scenarios across diverse language families with varying morphological and typological complexity. Baselines used symmetric BPE (\textit{4K\_4K}, \textit{32K\_32K}), while asymmetric settings (\textit{8K\_500}, \textit{16K\_500}) derive from English-Hindi optimal configurations at 0.1M sentence pairs. For evaluating we use the FLORES test set, except English$\leftrightarrow$Inuktitut tested on \citet{joanis-etal-2020-nunavut} (Appendix~\ref{subsec:validAndTest}).

\begin{figure}
\centering
\includegraphics[width=0.48\textwidth]{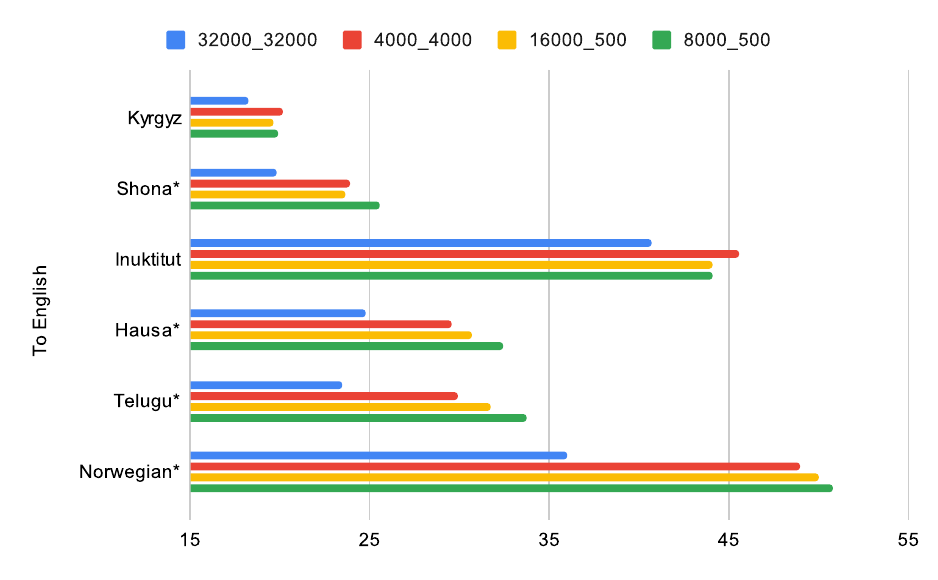}
\caption{CHRF++ scores improvement with asymmetrical over symmetrical BPE for English to \textit{L} Languages}\label{fig:othersEngTo}
\end{figure}

\begin{figure}
\centering
\includegraphics[width=0.48\textwidth]{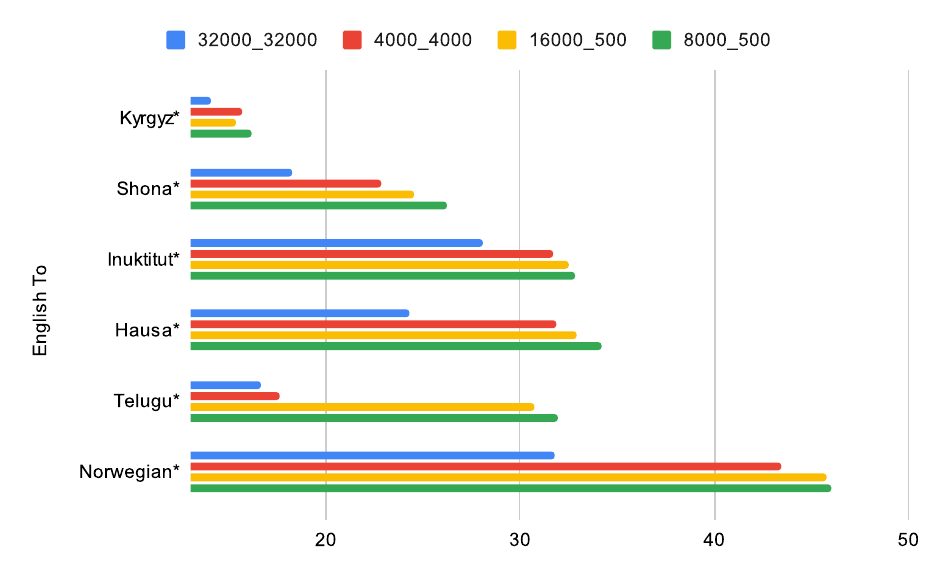}
\caption{CHRF++ scores improvement with asymmetrical over symmetrical BPE from \textit{L} Languages to English}\label{fig:others2Eng}
\end{figure}

Experiments are repeated three times for reproducibility (sampling, BPE training, model training). Figures~\ref{fig:othersEngTo} and~\ref{fig:others2Eng} compare average asymmetric and symmetric BPE results for translations to and from English. Asymmetric BPE significantly improves four of six \textit{L-to-English} systems and all \textit{English-to-L} systems ($p<0.05$, indicated by *), underscoring the benefits of asymmetric BPE and the need to explore beyond conventional settings for low-resource pairs.


\section{Conclusion}
\label{sec:conclusion}
In-depth examination of BPE configurations across diverse language pairs and differing dataset sizes reveals that typical configurations (\textit{n\_n}) do not always produce optimal results. As referenced in Section \ref{sec:relatedWork}, in low-resource settings, systems benefit from using symmetric \textit{n} NMO configurations when \textit{n} is significantly smaller than \textit{32K}; our experiments with asymmetric BPE \textit{n}\_\textit{m} show that further improvement in translation performance is possible, under low-resource conditions, when  \textit{n} >> \textit{m} where \textit{n}, \textit{m} represent NMOs for source and target respectively. 
This study highlights the need to go beyond default segmentation in machine translation, especially for low-resource languages. While symmetric BPE configurations may suffice with medium to large datasets, their effectiveness drops in low-resource settings. Using asymmetric BPE—with a higher number of merge operations for the source language and fewer for the target—yields significant translation quality gains. These configurations consistently outperform across varied language families and morphological complexities, underscoring the importance of tailored segmentation for optimizing low-resource translation.

\section*{Limitation}
This study is limited by the computational cost of exhaustively analysing all BPE configurations for each language pair and by its focus only on bilingual encoder–decoder NMT. However, the results show that certain configuration ranges consistently improve translation quality in low-resource settings, substantially reducing the search space. These findings suggest promising extensions to multilingual models, potentially combined with fairness-aware tokenisation such as Parity-Aware BPE \cite{foroutan2025parityawarebytepairencodingimproving} to deliver both performance gains and balanced vocabulary distribution.

\appendix

\section{Appendix}
\label{sec:appendix}
\subsection{English–Hindi Training Data Statistics}
\label{subsec:enhi_stats}
\begin{table*}[h!]
\centering
\resizebox{\width}{!}{%
\begin{tabular}{|l|r|r|r|r|r|r|}
\hline
\textbf{Length Range} & \textbf{\# of Lines} & \textbf{\% of Total} & \textbf{4M} & \textbf{1M} & \textbf{0.5M} & \textbf{0.1M} \\ \hline
1 to 10   & 2,792,334 & 34.13\% & 1,365,200 & 341,300 & 170,650 & 34,130 \\ \hline
11 to 15  & 1,655,162 & 20.23\% & 809,200   & 202,300 & 101,150 & 20,230 \\ \hline
16 to 20  & 1,150,396 & 14.06\% & 562,400   & 140,600 & 70,300  & 14,060 \\ \hline
21 to 25  & 854,091   & 10.44\% & 417,600   & 104,400 & 52,200  & 10,440 \\ \hline
31 to 35  & 420,583   & 5.14\%  & 205,600   & 51,400  & 25,700  & 5,140  \\ \hline
36 to 40  & 275,774   & 3.37\%  & 134,800   & 33,700  & 16,850  & 3,370  \\ \hline
$\geq$ 41 & 414,926   & 5.07\%  & 202,800   & 50,700  & 25,350  & 5,070  \\ \hline
\textbf{Total} & \textbf{8,180,584} &         & \textbf{3,999,600} & \textbf{999,900} & \textbf{499,950} & \textbf{99,990} \\ \hline
\end{tabular}
}
\caption{Distribution of English–Hindi sentence pairs sampled from Samanantar across sentence length bins and different dataset sizes.}
\label{tab:eh-sampling}
\end{table*}
We use an 8-million-sentence English–Hindi corpus from the Samanantar dataset and execute stratified random sampling across sentence length bins to simulate different resource availability levels. Table \ref{tab:eh-sampling} summarises the statistics for sentence pairs corresponding to each level of resource availability.


\subsection{Hyperparameters for Training Transformer Model}
\label{subsec:hyperparameters}
\begin{table}[h!]
\centering
\resizebox{\columnwidth}{!}{%
\begin{tabular}{|c|c|}
\hline
\textbf{Parameter} & \textbf{Value} \\ \hline
\texttt{arch} & transformer \\ \hline
\texttt{optimizer} & adam \\ \hline
\texttt{adam-betas} & (0.9, 0.98) \\ \hline
\texttt{clip-norm} & 0.0 \\ \hline
\texttt{lr} & 5e-4 \\ \hline
\texttt{lr-scheduler} & inverse\_sqrt \\ \hline
\texttt{warmup-updates} & 4000 \\ \hline
\texttt{warmup-init-lr} & 1e-07 \\ \hline
\texttt{dropout} & 0.3 \\ \hline
\texttt{attention-dropout} & 0.1 \\ \hline
\texttt{activation-dropout} & 0.1 \\ \hline
\texttt{weight-decay} & 0.0001 \\ \hline
\texttt{criterion} & label\_smoothed\_cross\_entropy \\ \hline
\texttt{label-smoothing} & 0.1 \\ \hline
\texttt{max-tokens} & 6000 \\ \hline
\texttt{max-update} & 300000 \\ \hline
\texttt{patience} & 20 \\ \hline
\texttt{update-freq} & 10 \\ \hline
\end{tabular}}
\caption{Training hyperparameters used across all experiments.}
\label{tab:hyperparams}
\end{table}
We followed the official Fairseq tutorial instructions for preprocessing, training, and translation\footnote{\url{https://fairseq.readthedocs.io/en/latest/getting_started.html}}, and customised the parameters given in Table \ref{tab:hyperparams} with respective values for all experiments.


\subsection{Performance of all systems for English $\leftrightarrow$ Hindi for all dataset scenarios}
\label{subsec:allSystemsENHI}
\begin{figure*}[htbp]
\centering

\subfloat[0.05 Million English to Hindi]{\includegraphics[width=0.48\textwidth]{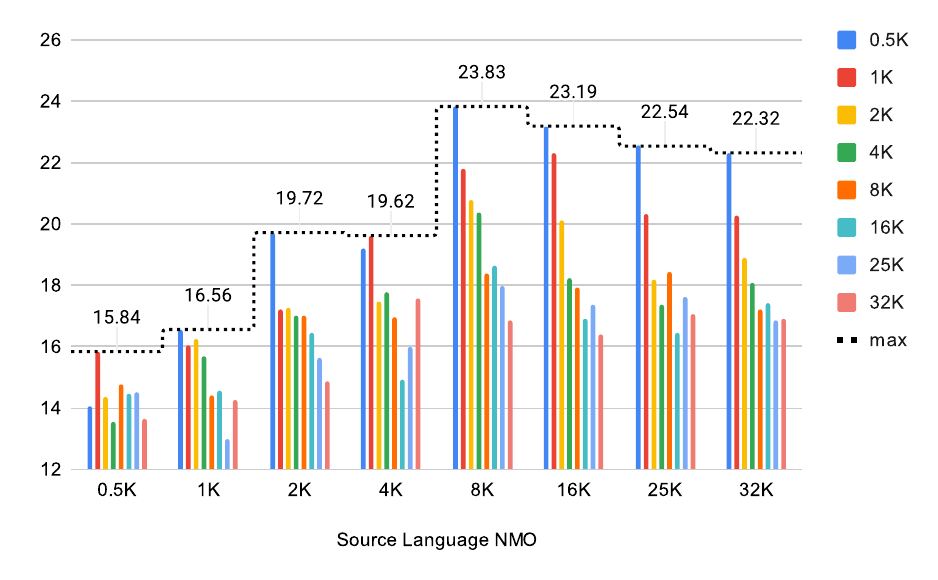}}\hfill
\subfloat[0.05 Million Hindi to English]{\includegraphics[width=0.48\textwidth]{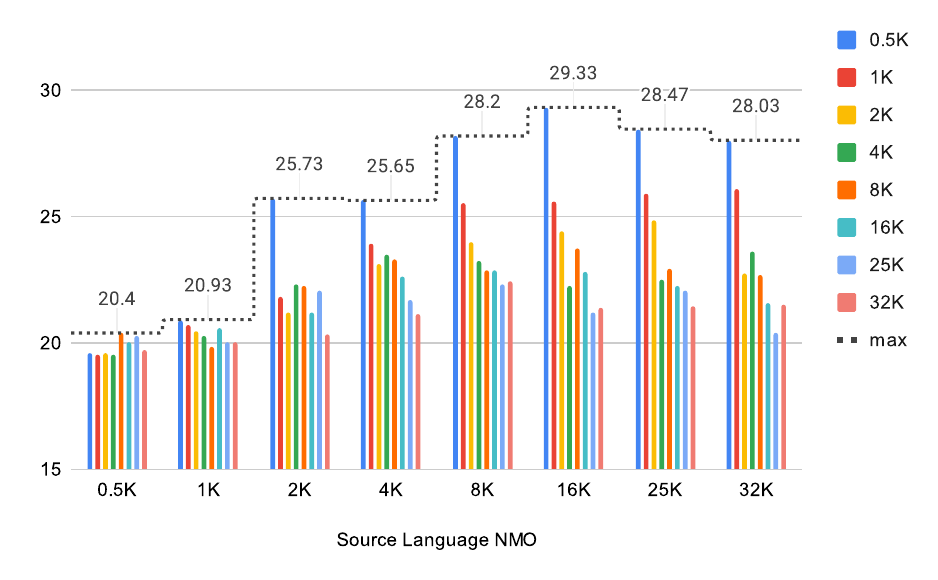}}\hfill

\subfloat[0.1 Million English to Hindi]{\includegraphics[width=0.50\textwidth]{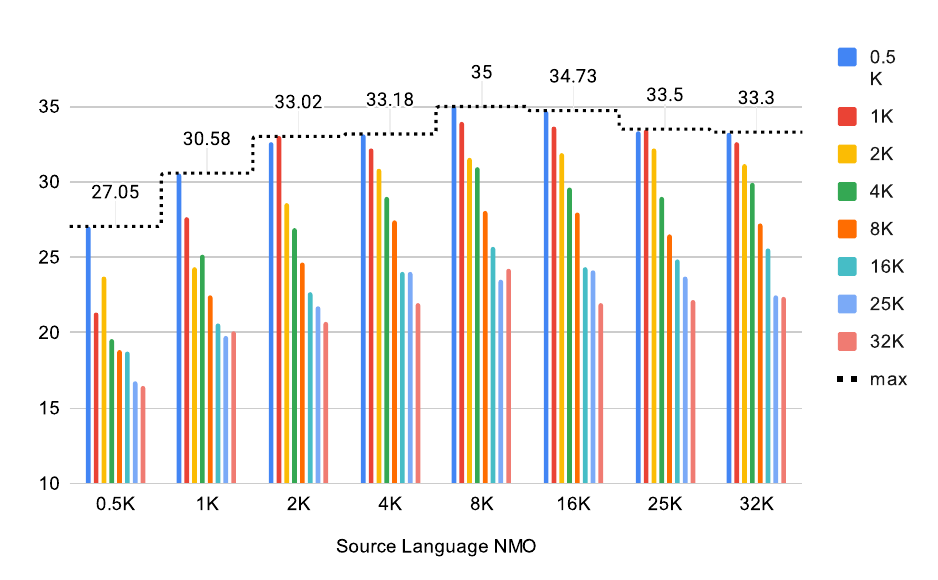}}\hfill
\subfloat[0.1 Million Hindi to English]{\includegraphics[width=0.50\textwidth]{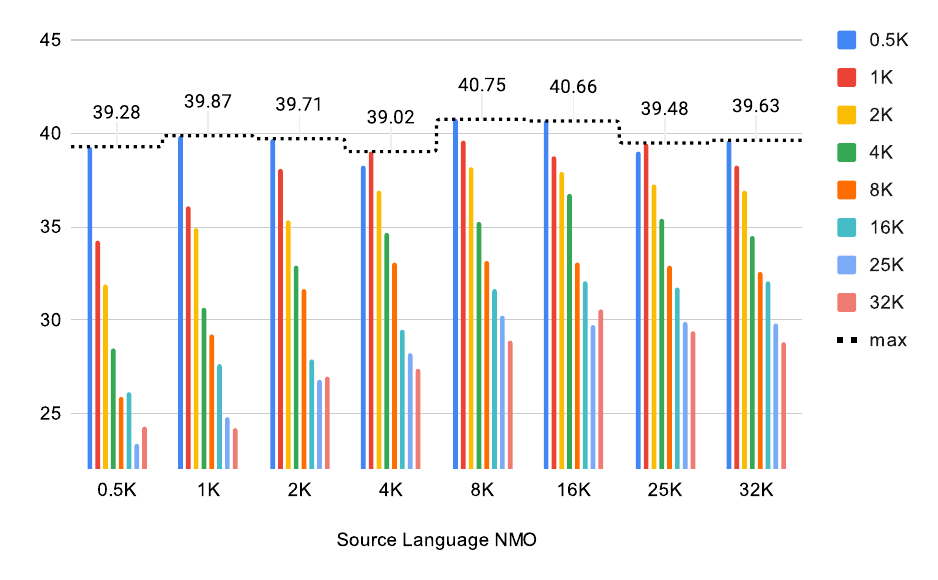}}\hfill

\subfloat[0.5 Million English to Hindi]{\includegraphics[width=0.48\textwidth]{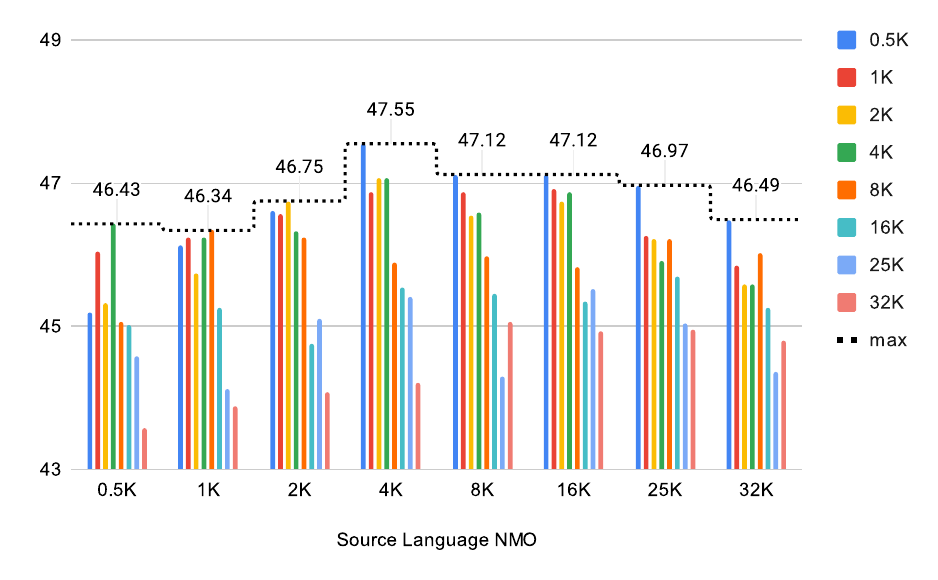}}\hfill
\subfloat[0.5 Million Hindi to English]{\includegraphics[width=0.48\textwidth]{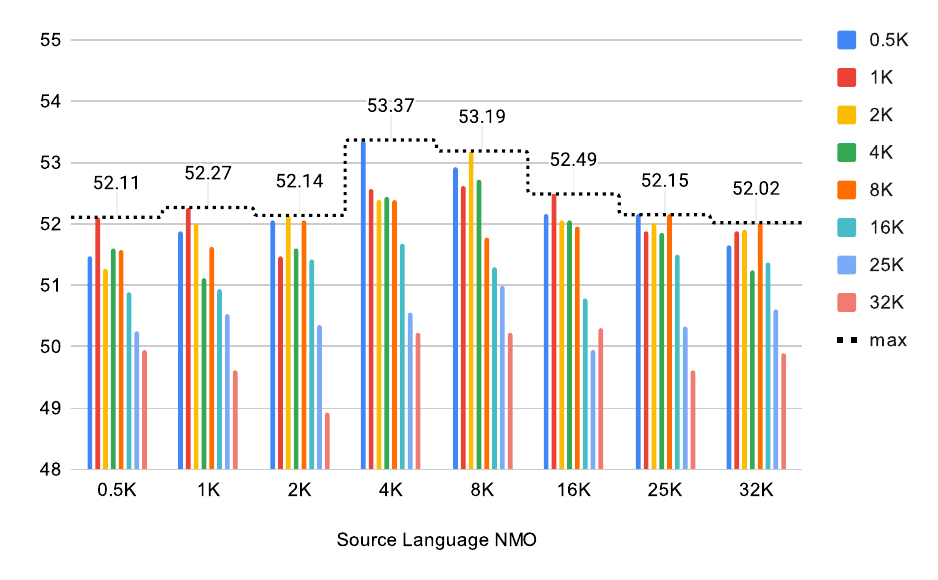}}\hfill

\caption{Evaluation of English $\leftrightarrow$ Hindi MT Systems for 0.05M, 0.1M and 0.5M dataset sizes on FLORES, x-axis is source NMO and y-axis is CHRF++ scores}\label{fig:ENHILow}

\end{figure*}

\begin{figure*}[htbp]
\centering

\subfloat[1 Million English to Hindi]{\includegraphics[width=0.50\textwidth]{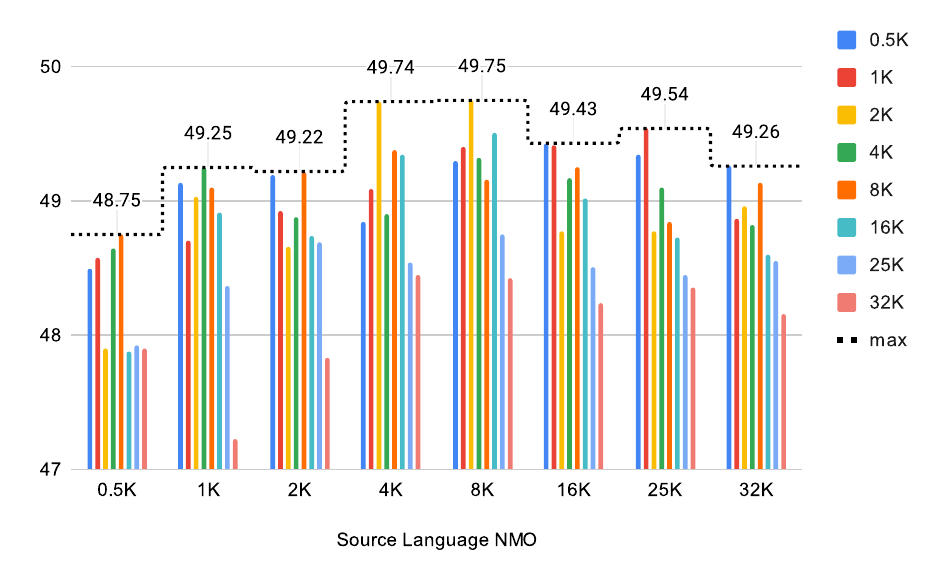}}\hfill
\subfloat[1 Million Hindi to English]{\includegraphics[width=0.50\textwidth]{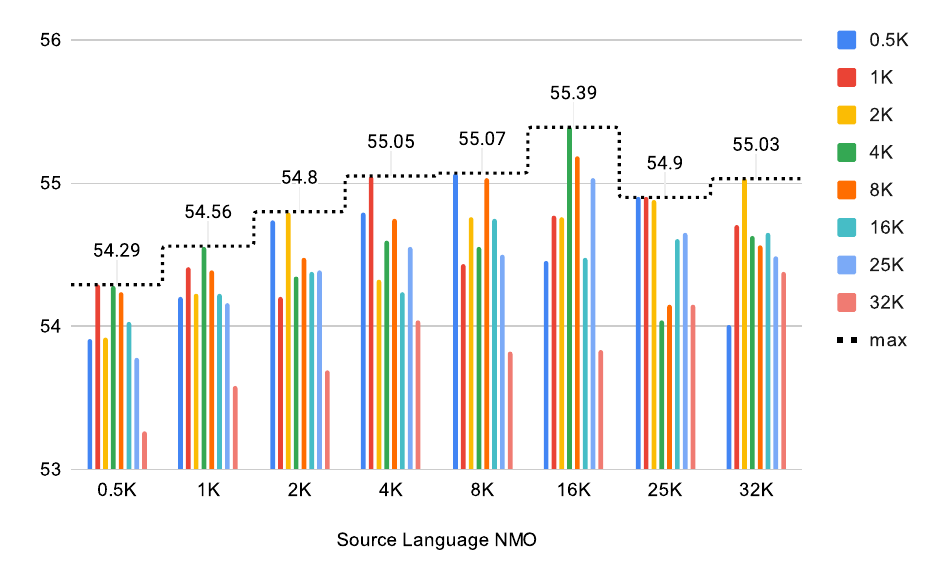}}\hfill

\subfloat[4 Million English to Hindi]{\includegraphics[width=0.50\textwidth]{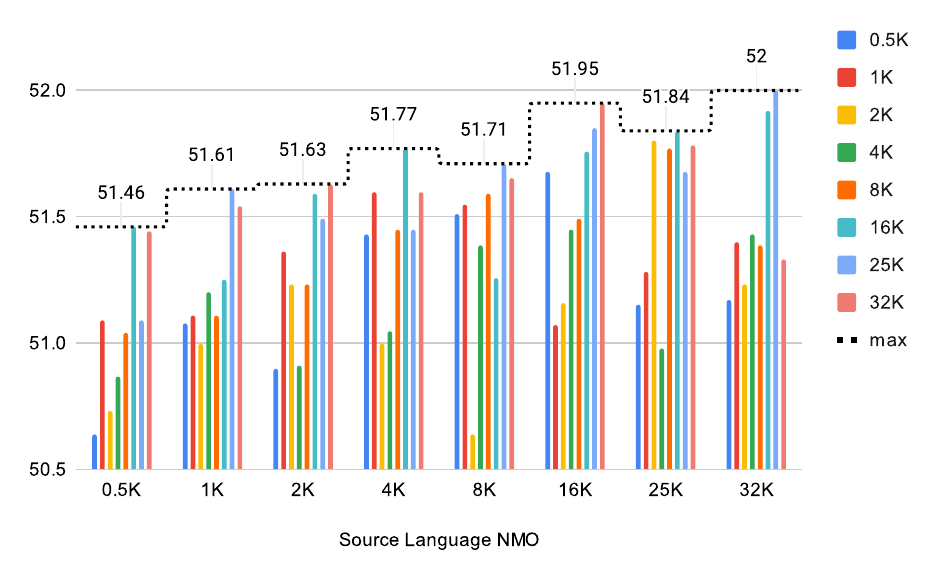}}\hfill
\subfloat[4 Million Hindi to English]{\includegraphics[width=0.50\textwidth]{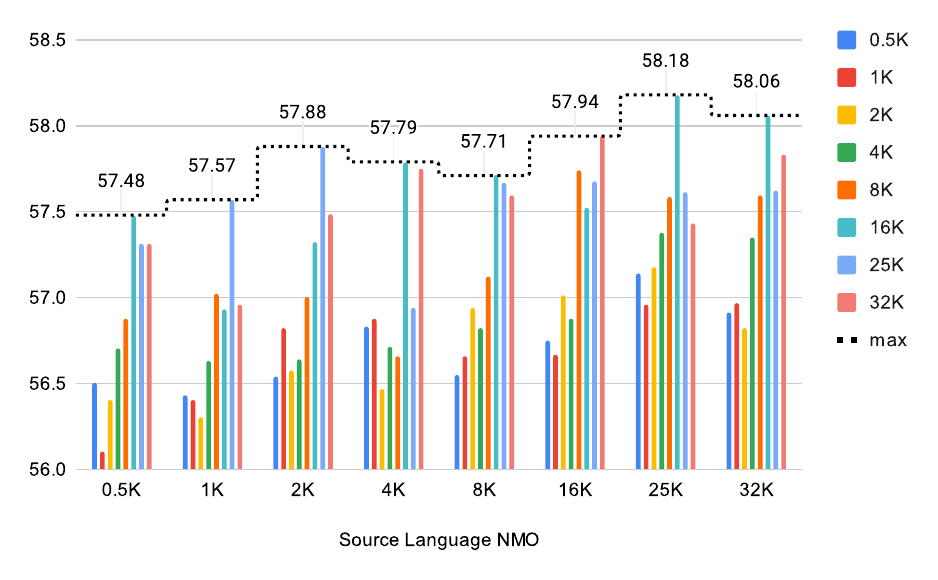}}\hfill

\subfloat[8 Million English to Hindi]{\includegraphics[width=0.50\textwidth]{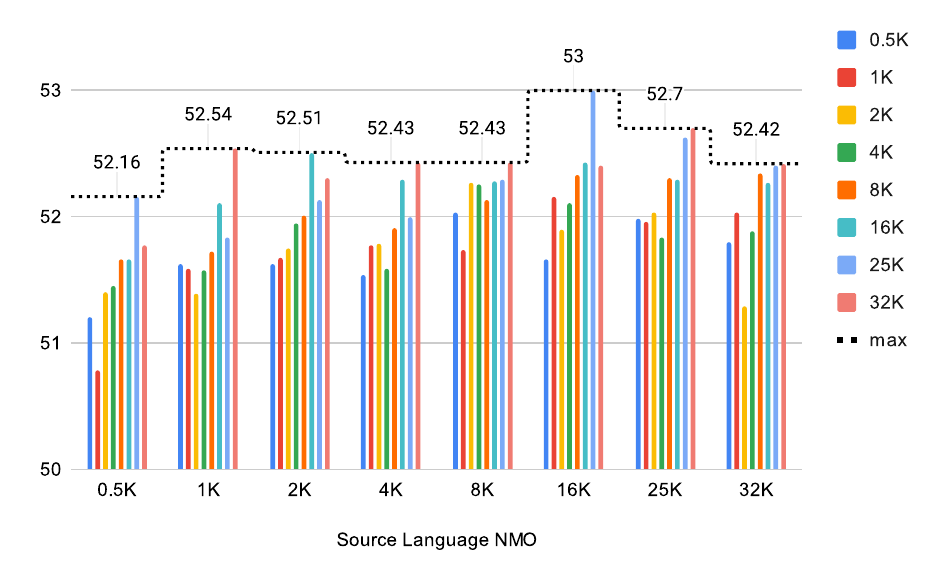}}\hfill
\subfloat[8 Million Hindi to English]{\includegraphics[width=0.50\textwidth]{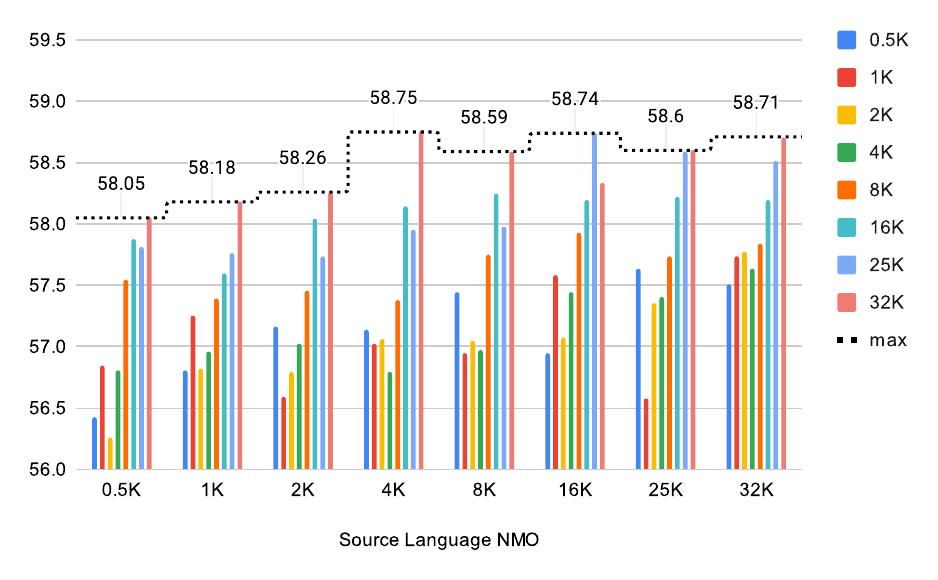}}\hfill
\caption{Evaluation of English $\leftrightarrow$ Hindi MT Systems for 1M, 4M and 8M dataset sizes on FLORES, x-axis is source NMO and y-axis is CHRF++ scores}\label{fig:ENHIHigh}
\end{figure*}

Figures \ref{fig:ENHILow} present the performance of all configurations for English $\leftrightarrow$ Hindi systems in a low resource scenario (for data set sizes of 0.05M, 0.1M and 0.5M). And Figures \ref{fig:ENHIHigh} show the performance of all configurations on 1M, 4M and 8M dataset sizes. Each subgraph represents performance on a particular dataset size, with the x-axis being the source NMO. The black stepped dotted lines indicate the maximum CHRF++ score for each dataset size considering for each source NMOs. 
In figure \ref{fig:ENHILow} for low-resource environments (0.05M, 0.1M and 0.5M) systems, as noted by \cite{ding-etal-2019-call}, the use of symmetric BPE configuration with lower NMOs improves performance over high NMOs. However, the best results are achieved using asymmetric BPE configurations when the source has a higher NMO than the target. We see a maximum performance gain when the source NMO is very high and the target NMO very low (we see consistent performance with the target NMO = \textit{500}). Conversely, when the target's NMO is greater than that of the source, performance declines, like for the Hindi to English 0.1M dataset, performance of \textit{500}\_\textit{25K} and \textit{500}\_\textit{32K} was worse than symmetric BPE configurations.

\subsection{Performance of Hindi-To-English Selected Configurations on Domain Test set}
\label{subsec:hin2eng_domain}
Table~\ref{tab:domain_hi2en} shows the performance of the Highest and Lowest performing asymmetric BPE systems with baseline systems for Hindi-To-English systems. Like in English to Hindi systems, we see significant improvement when using asymmetric BPE configurations in low-resource settings.
\begin{table*}
\centering
\resizebox{0.8\textwidth}{!}{%
\begin{tblr}{
  cells = {c},
  cell{1}{2} = {c=4}{},
  cell{1}{6} = {c=4}{},
  cell{1}{10} = {c=4}{},
  cell{8}{2} = {c=4}{},
  cell{8}{6} = {c=4}{},
  cell{8}{10} = {c=4}{},
  vline{2-3,7} = {1,8}{},
  vline{2,6,10} = {2-7,9-14}{},
  hline{1-3,8-10,15} = {-}{},
}
Dataset
  Size   & 0.05M &     &                 &                 & 0.1M &     &                 &                 & 0.5M &     &                 &       \\
Performance Tier & src   & tgt & AI              & CH              & src  & tgt & AI              & CH              & src  & tgt & AI              & CH    \\
Low A            & 500    & 1K & 20.87            & 19.64          & 500  & 25K & 25.22           & 23.56           & 2K   & 32K & 57.8            & 50.82 \\
Low B            & 500    & 2K & 19.71           & 18.46           & 1K   & 32K & 26.65           & 24.61           & 25K  & 32K & 59.35           & 52.27 \\
Baseline         & 4K    & 4K  & 24.61           & 22.92           & 500  & 500 & 47.55           & 41.21           & 4K   & 4K  & 63.7            & 56.61 \\
High B           & 25K   & 500 & \textbf{31.22*} & \textbf{28.17*} & 16K  & 500 & \textbf{50.36*} & \textbf{42.12*} & 8K   & 2K  & 64.07           & 56.61 \\
High A           & 16K   & 500 & \textbf{32.74*} & \textbf{29.78*} & 8K   & 500 & \textbf{50.41*} & \textbf{42.41*} & 4K   & 500 & \textbf{64.29*} & 56.84 \\
Dataset Size     & 1M    &     &                 &                 & 4M   &     &                 &                 & 8M   &     &                 &       \\
Performance Tier & src   & tgt & AI              & CH              & src  & tgt & AI              & CH              & src  & tgt & AI              & CH    \\
Low A            & 500   & 32K & 63.75           & 57.23           & 500  & 1K  & 67.51           & 61.19           & 500  & 2K  & 68.02           & 61.3  \\
Low B            & 1K    & 32K & 64.33           & 57.13           & 1K   & 2K  & 67.86           & 61.55           & 500  & 500 & 68.12           & 61.24 \\
Baseline         & 8K    & 8K  & 65.52           & 59.18           & 32K  & 32K & 68.1            & 62.1            & 32K  & 32K & 69.74           & 63.24 \\
High B           & 16K   & 8K  & \textbf{66.07*} & 59.03           & 32K  & 16K & 68.08           & 61.94           & 16K  & 25K & 69.47           & 63.05 \\
High A           & 16K   & 4K  & 65.68           & 60.11           & 25K  & 16K & \textbf{69.32}  & 62.45           & 4K   & 32K & 69.68           & 63.18 
\end{tblr}}
\caption{Performance of the top 2 (High A and High B) and bottom 2 (Low A and Low B) systems with respective tokenisation configurations compared to the symmetric baseline for \textit{Hindi-to-English} systems across dataset sizes for \textbf{AI} and \textbf{CH Domains}. Bold scores indicate statistically significant improvements over the baseline ($p\textless{}0.05$); bold scores with an asterisk ($*$) indicate high significance ($p\textless{}0.01$)}
\label{tab:domain_hi2en}

\end{table*}
\subsection{Evaluation of English $\leftrightarrow$ Hindi systems on AI for all BPE Configurations}
\label{ch:appendixAI}
\begin{figure*}[!htbp]
\centering

\subfloat[0.05 Million English to Hindi]{\includegraphics[width=0.48\textwidth]{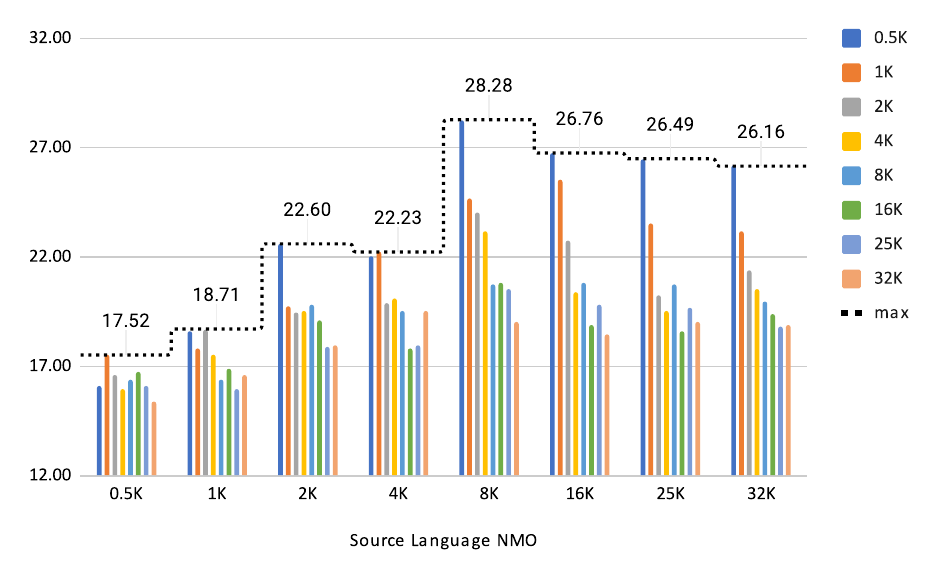}}\hfill
\subfloat[0.05 Million Hindi to English]{\includegraphics[width=0.48\textwidth]{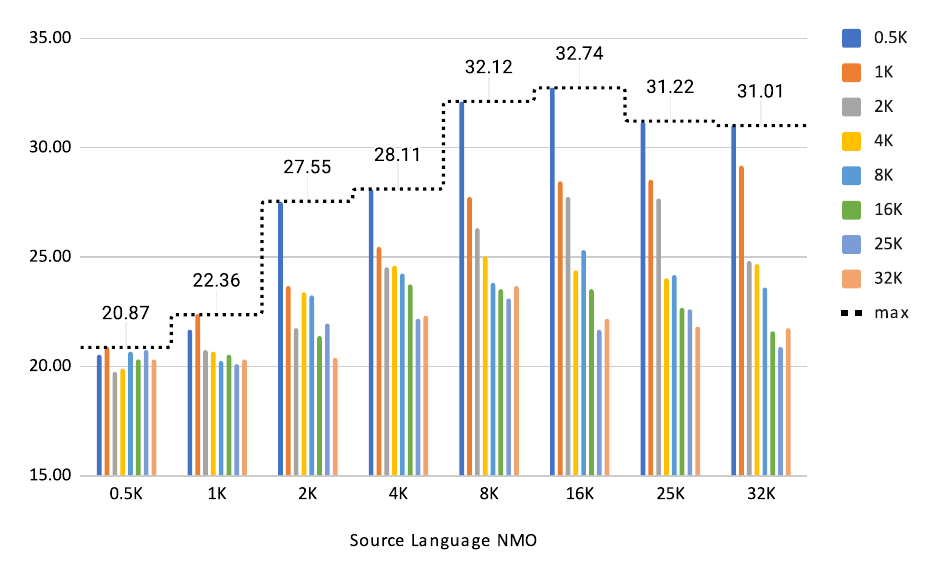}}\hfill

\subfloat[0.1 Million English to Hindi]{\includegraphics[width=0.50\textwidth]{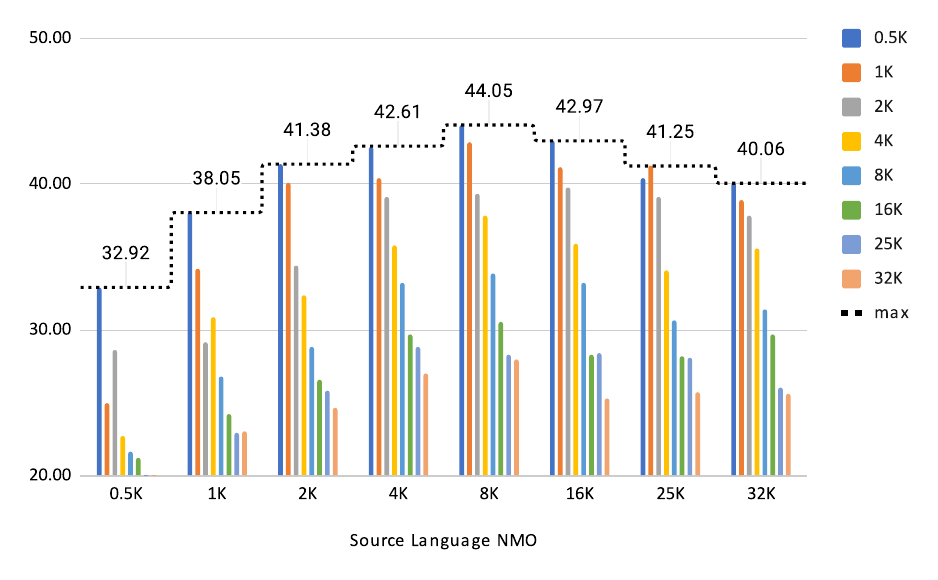}}\hfill
\subfloat[0.1 Million Hindi to English]{\includegraphics[width=0.50\textwidth]{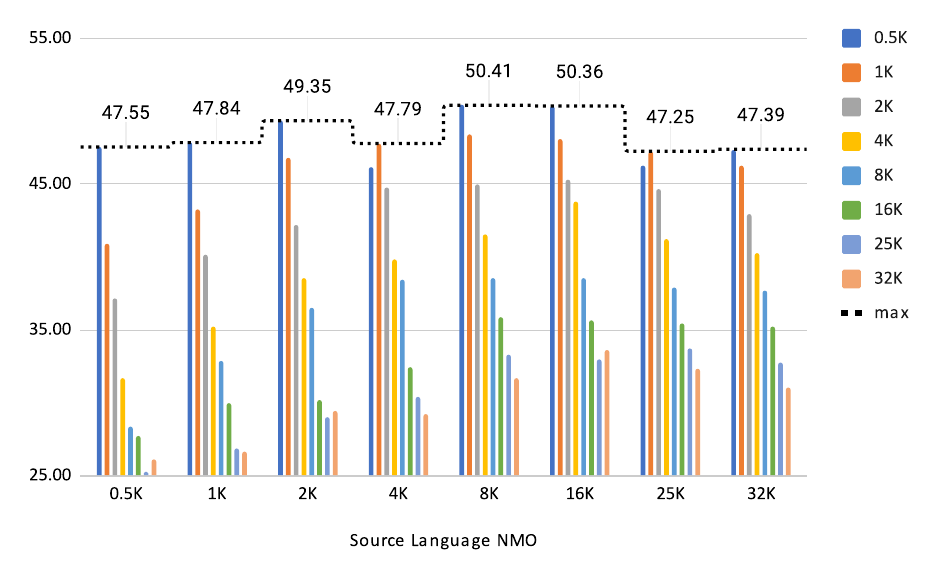}}\hfill

\subfloat[0.5 Million English to Hindi]{\includegraphics[width=0.48\textwidth]{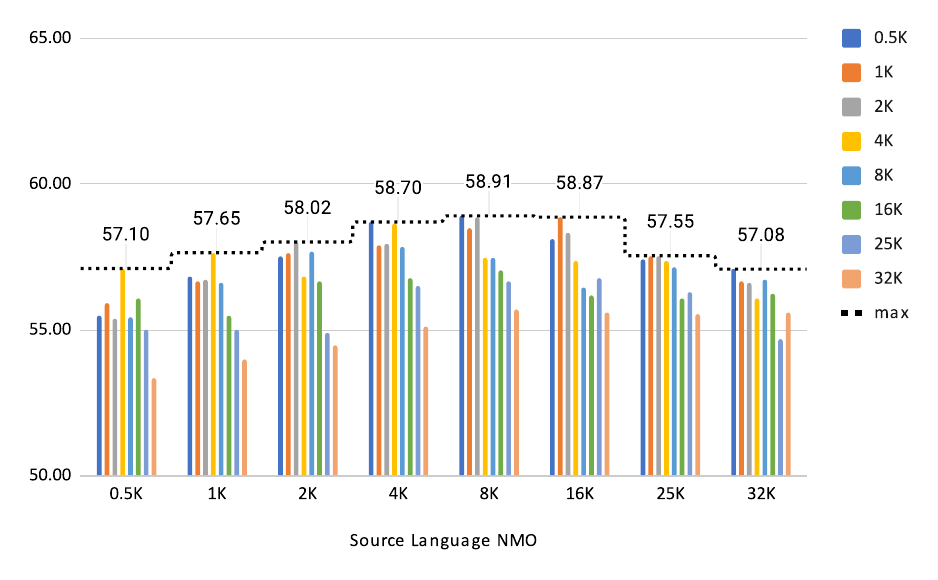}}\hfill
\subfloat[0.5 Million Hindi to English]{\includegraphics[width=0.48\textwidth]{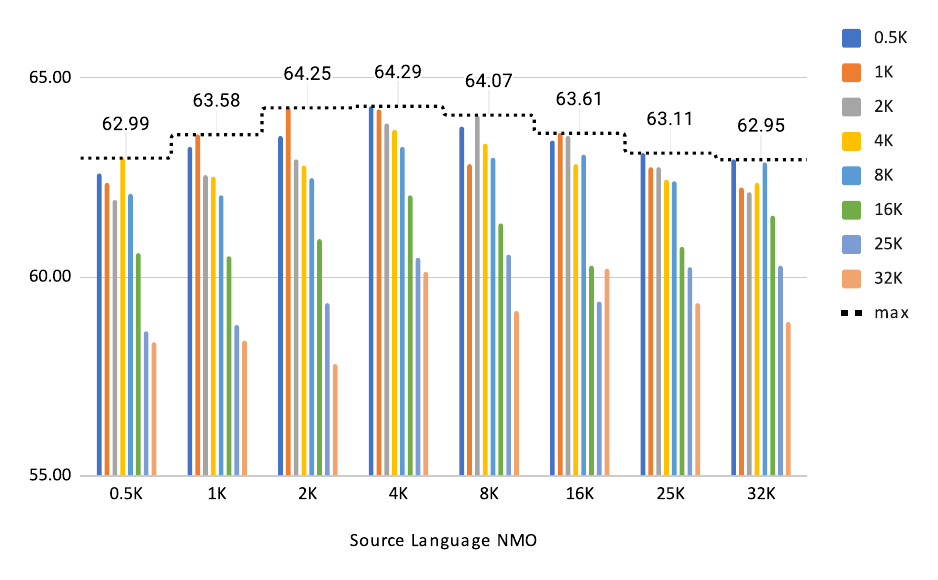}}\hfill

\caption{Evaluation of English $\leftrightarrow$ Hindi MT Systems for 0.05M, 0.1M and 0.5M dataset sizes on \textbf{AI}, x-axis is source NMO and y-axis is CHRF++ scores}\label{fig:ENHILowAI}

\end{figure*}

\begin{figure*}[htbp]
\centering

\subfloat[1 Million English to Hindi]{\includegraphics[width=0.50\textwidth]{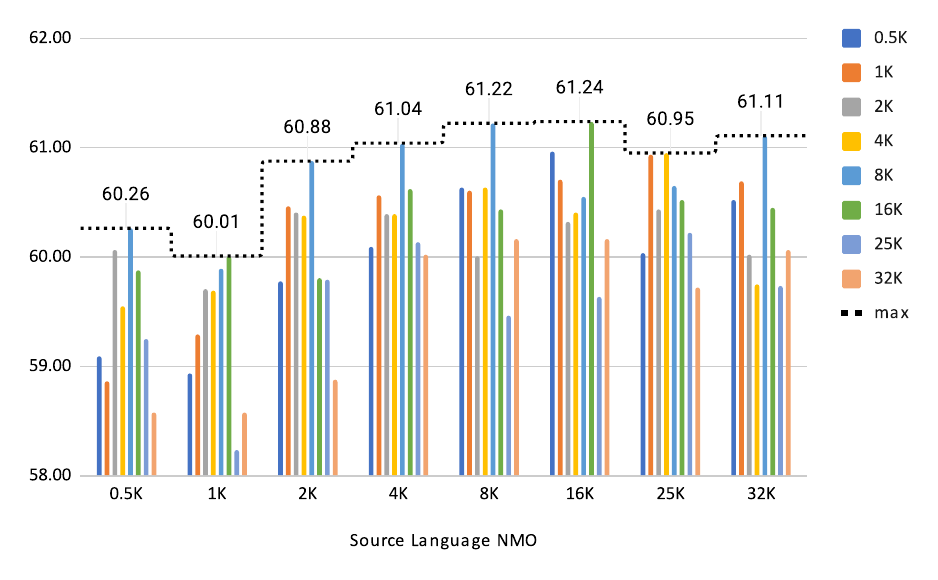}}\hfill
\subfloat[1 Million Hindi to English]{\includegraphics[width=0.50\textwidth]{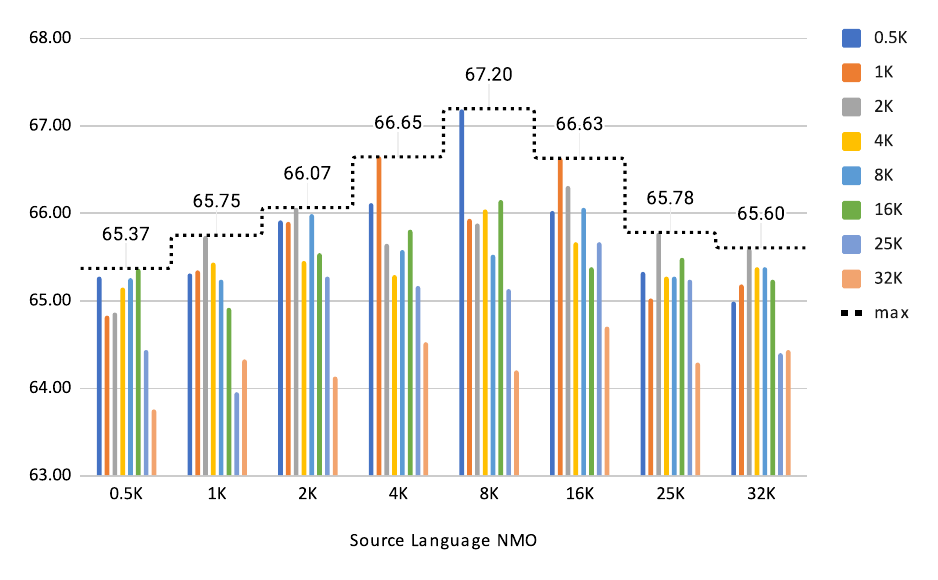}}\hfill

\subfloat[4 Million English to Hindi]{\includegraphics[width=0.50\textwidth]{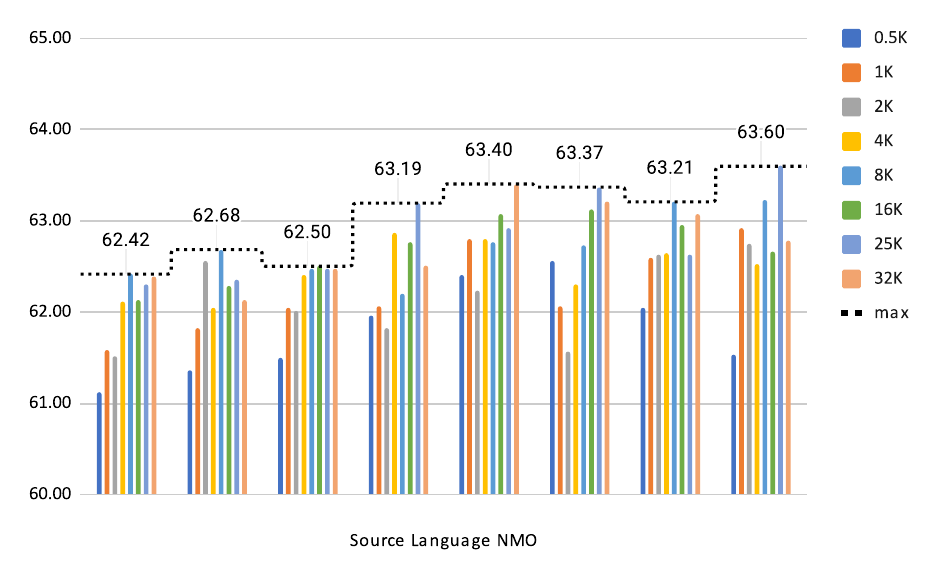}}\hfill
\subfloat[4 Million Hindi to English]{\includegraphics[width=0.50\textwidth]{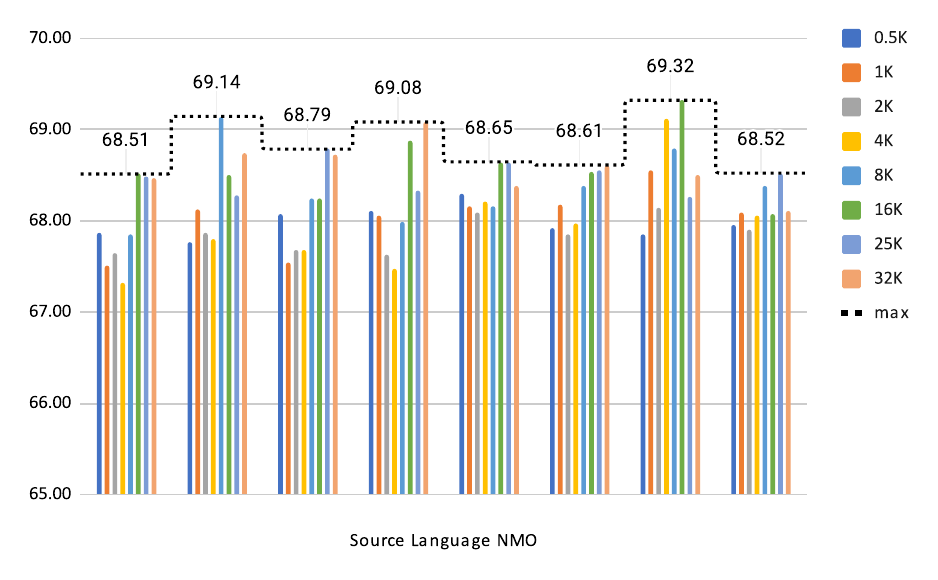}}\hfill

\subfloat[8 Million English to Hindi]{\includegraphics[width=0.50\textwidth]{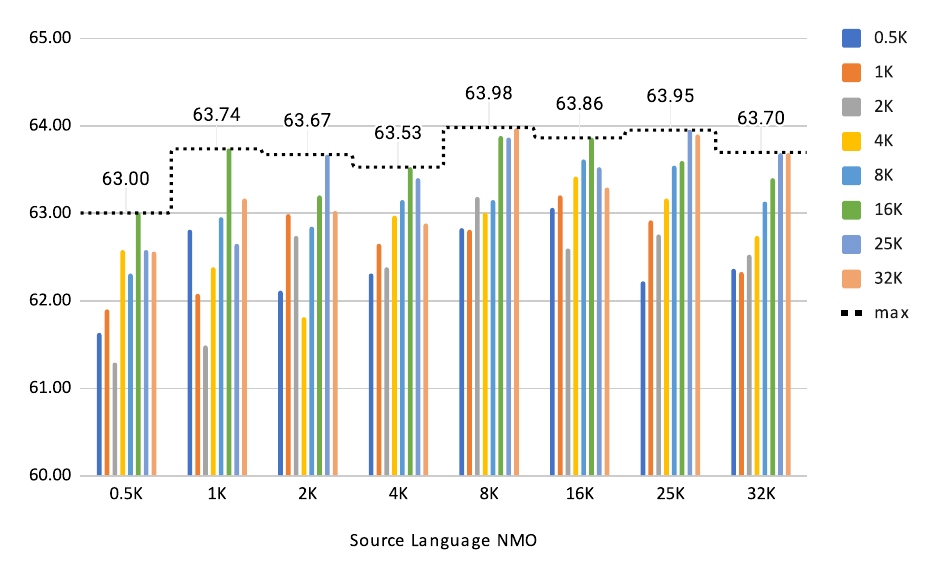}}\hfill
\subfloat[8 Million Hindi to English]{\includegraphics[width=0.50\textwidth]{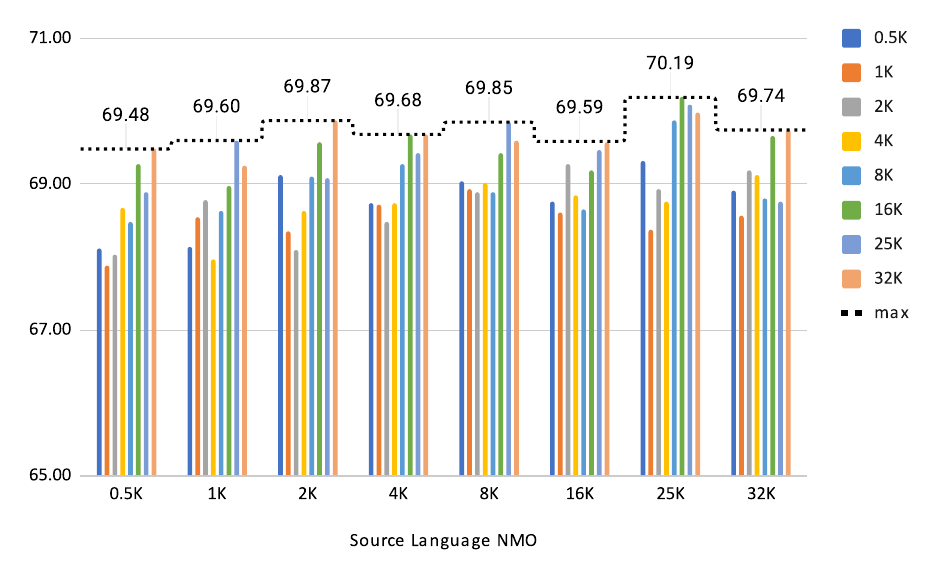}}\hfill
\caption{Evaluation of English $\leftrightarrow$ Hindi MT Systems for 1M, 4M and 8M dataset sizes on \textbf{AI}, x-axis is source NMO and y-axis is CHRF++ scores}\label{fig:ENHIHighAI}
\end{figure*}

Figures \ref{fig:ENHILowAI} and \ref{fig:ENHIHighAI} depict the performance of all configurations for English $\leftrightarrow$ Hindi systems during translations in the \textbf{AI} domain. A similar performance pattern appears across configurations here, as observed with the FLORES test set (see Appendix \ref{subsec:allSystemsENHI}).

\subsection{Evaluation of English $\leftrightarrow$ Hindi systems on Chemistry for all BPE Configurations}
\label{ch:appendixCH}
\begin{figure*}[htbp]
\centering
\subfloat[0.05 Million English to Hindi]{\includegraphics[width=0.48\textwidth]{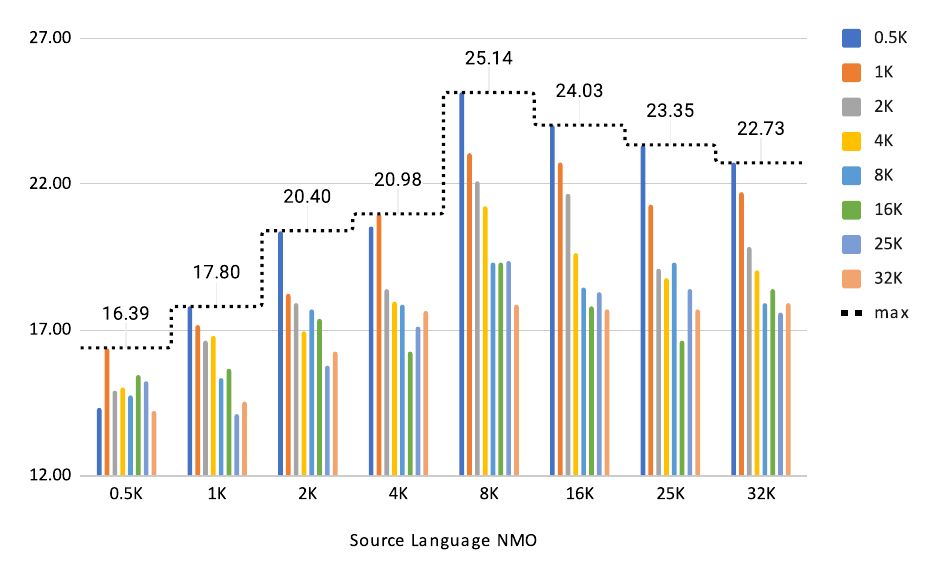}}\hfill
\subfloat[0.05 Million Hindi to English]{\includegraphics[width=0.48\textwidth]{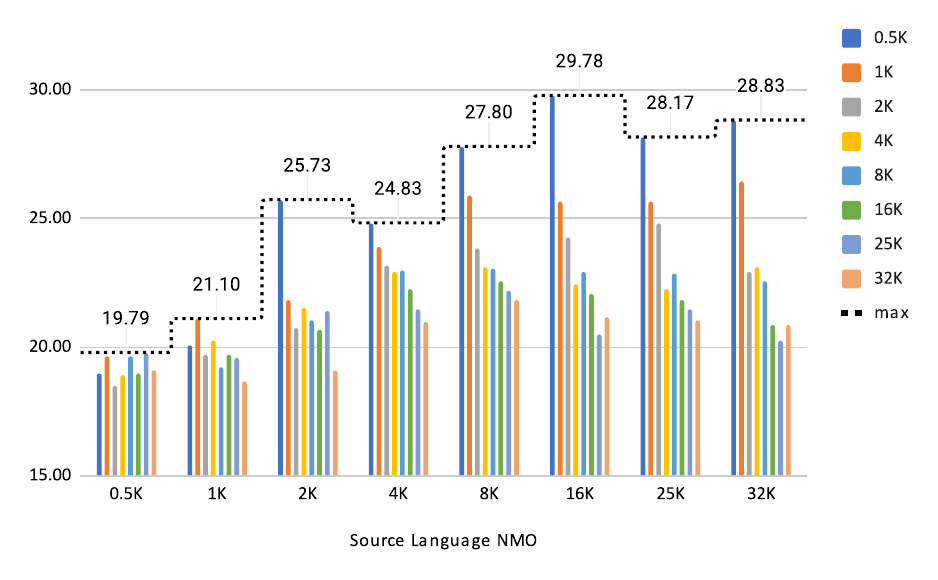}}\hfill

\subfloat[0.1 Million English to Hindi]{\includegraphics[width=0.50\textwidth]{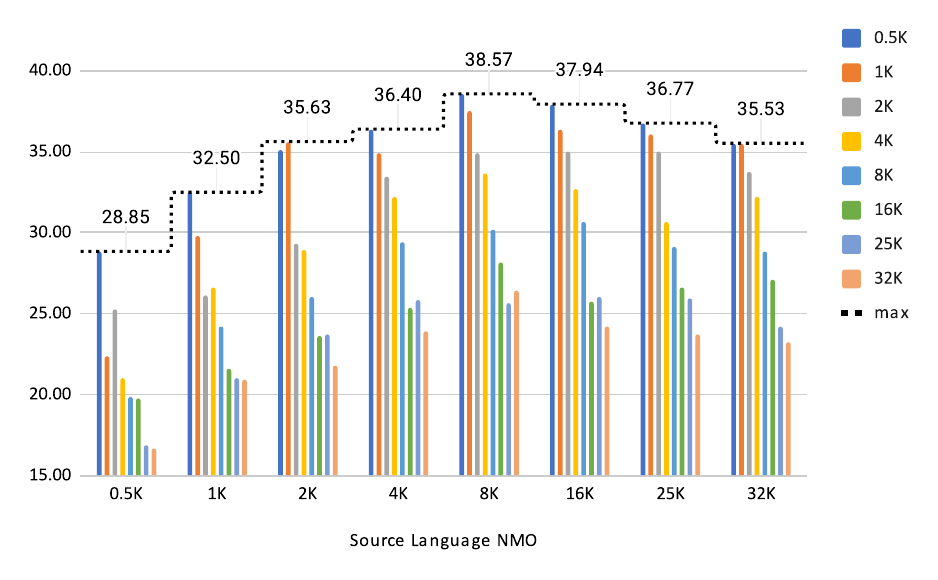}}\hfill
\subfloat[0.1 Million Hindi to English]{\includegraphics[width=0.50\textwidth]{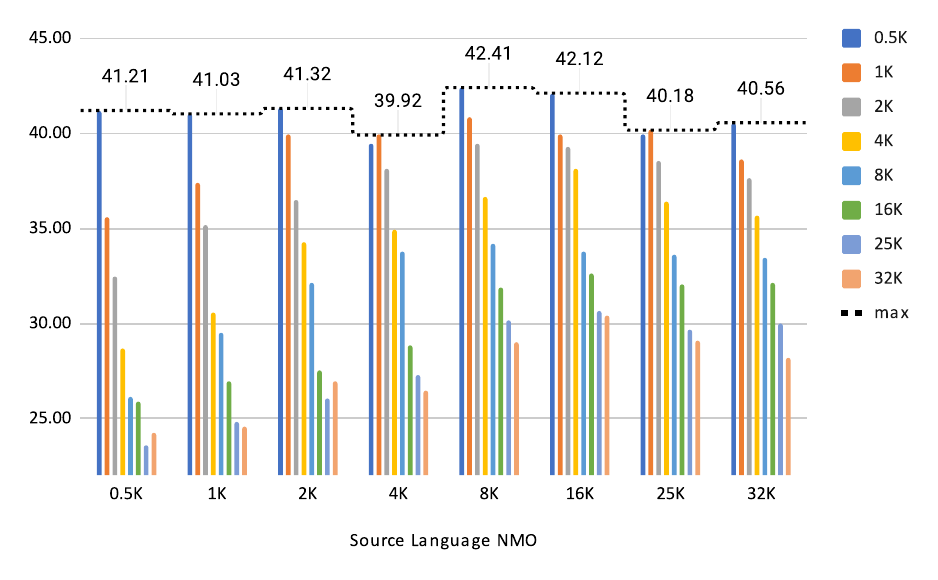}}\hfill

\subfloat[0.5 Million English to Hindi]{\includegraphics[width=0.48\textwidth]{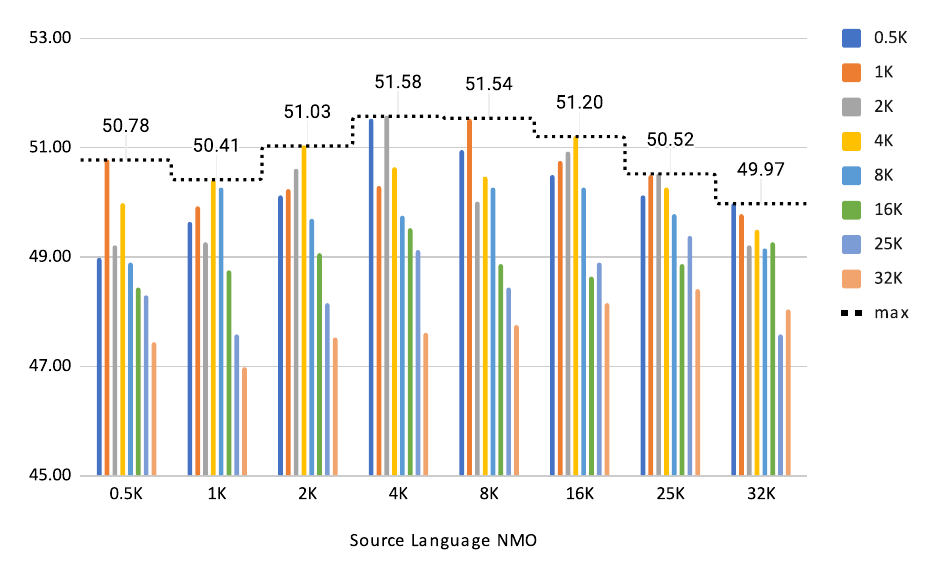}}\hfill
\subfloat[0.5 Million Hindi to English]{\includegraphics[width=0.48\textwidth]{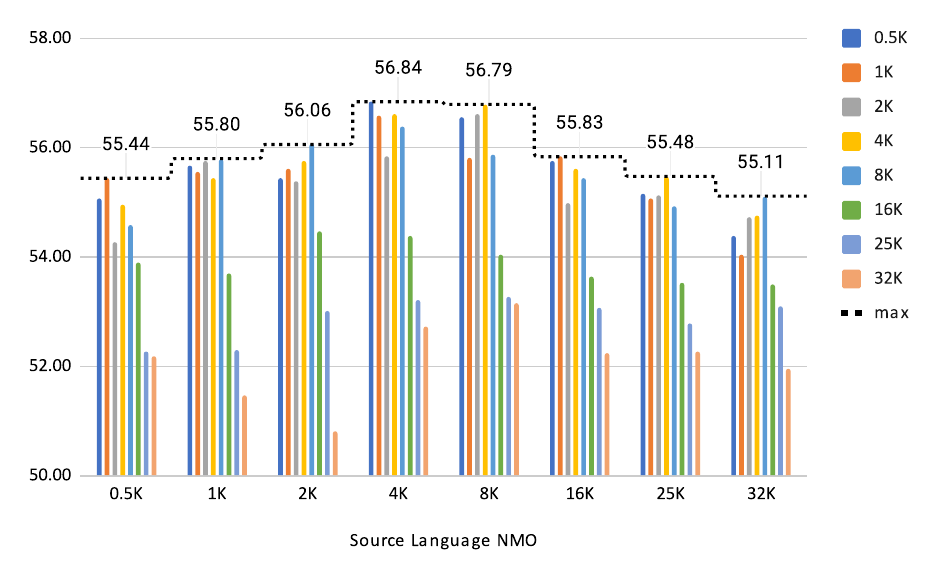}}\hfill

\caption{Evaluation of English $\leftrightarrow$ Hindi MT Systems for 0.05M, 0.1M and 0.5M dataset sizes on \textbf{CH}, x-axis is source NMO and y-axis is CHRF++ scores}\label{fig:ENHILowCH}

\end{figure*}

\begin{figure*}[htbp]
\centering
\subfloat[1 Million English to Hindi]{\includegraphics[width=0.50\textwidth]{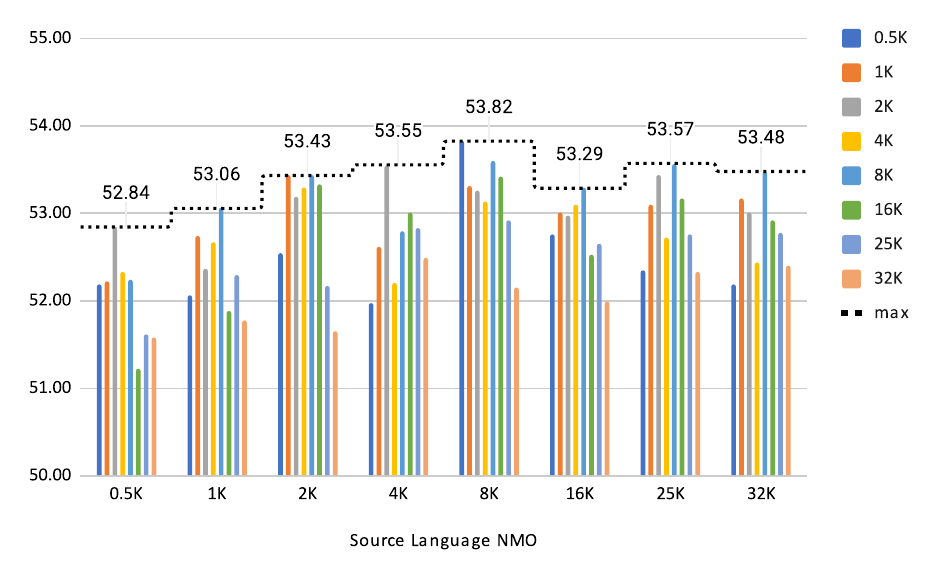}}\hfill
\subfloat[1 Million Hindi to English]{\includegraphics[width=0.50\textwidth]{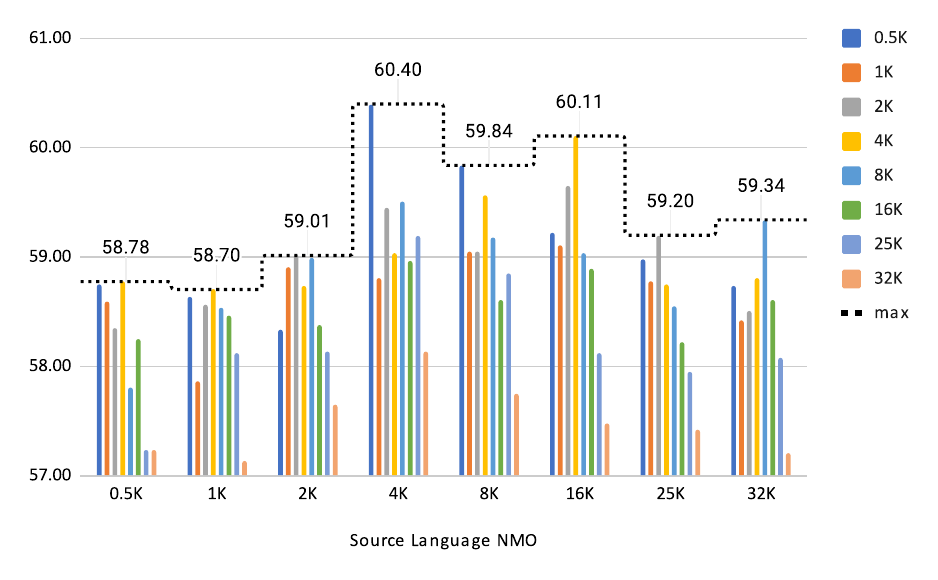}}\hfill

\subfloat[4 Million English to Hindi]{\includegraphics[width=0.50\textwidth]{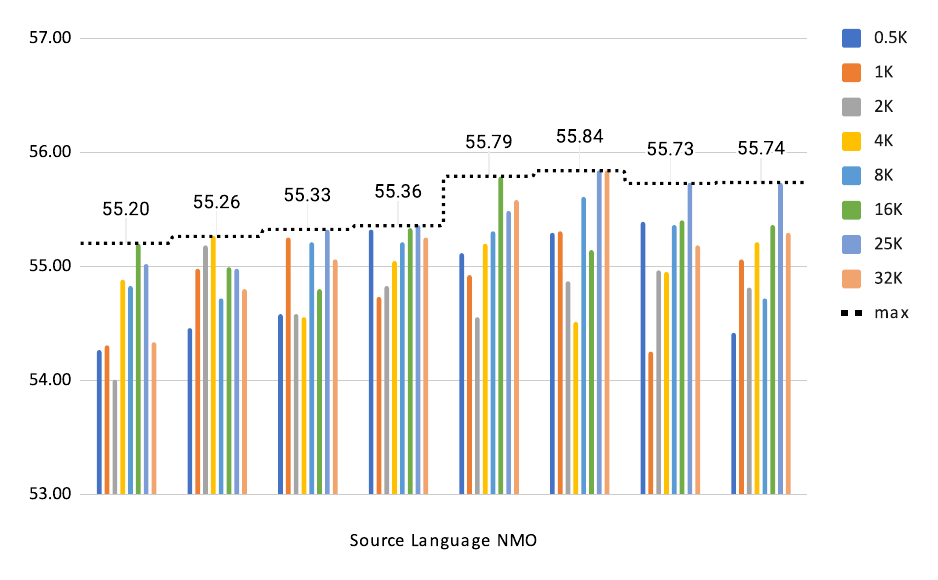}}\hfill
\subfloat[4 Million Hindi to English]{\includegraphics[width=0.50\textwidth]{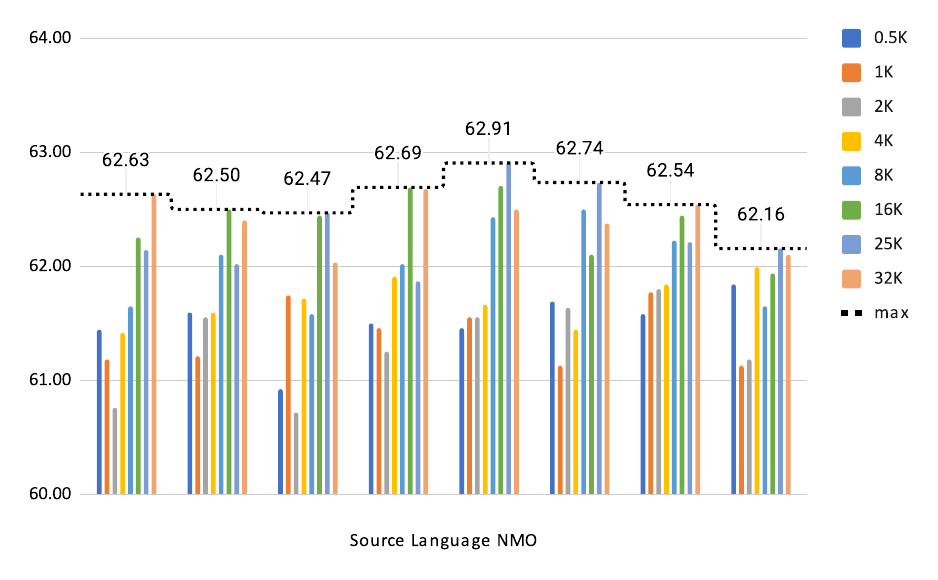}}\hfill

\subfloat[8 Million English to Hindi]{\includegraphics[width=0.50\textwidth]{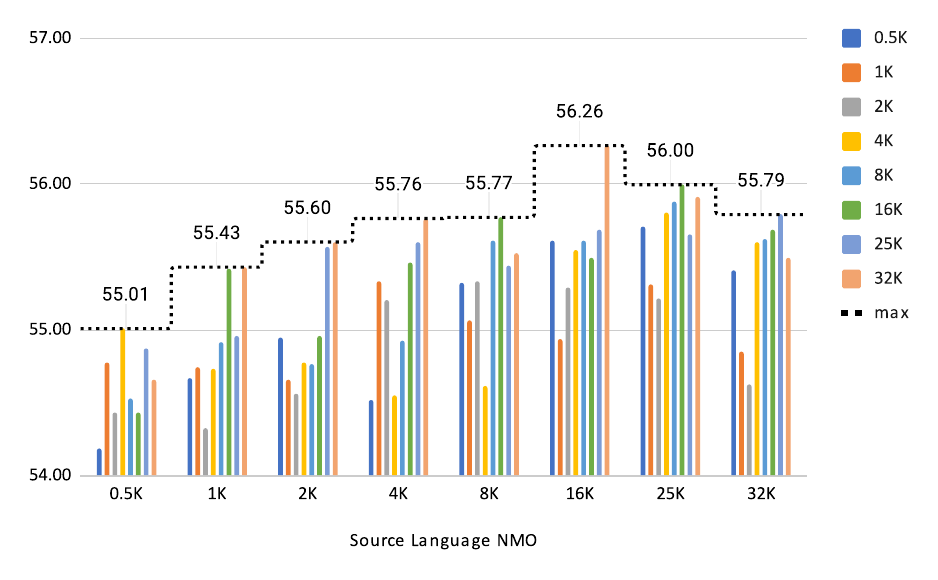}}\hfill
\subfloat[8 Million Hindi to English]{\includegraphics[width=0.50\textwidth]{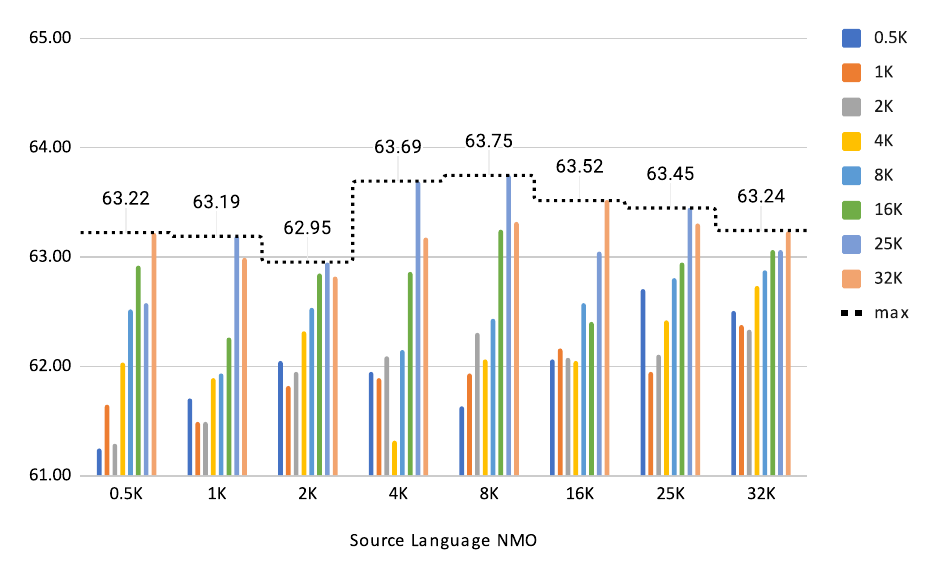}}\hfill
\caption{Evaluation of English $\leftrightarrow$ Hindi MT Systems for 1M, 4M and 8M dataset sizes on \textbf{CH}, x-axis is source NMO and y-axis is CHRF++ scores}\label{fig:ENHIHighCH}
\end{figure*}

Figures \ref{fig:ENHILowCH} and \ref{fig:ENHIHighCH} depict the performance of all configurations for English $\leftrightarrow$ Hindi systems during translations in the \textbf{Chemistry} domain. A similar performance pattern appears across configurations here, as observed with the FLORES test set (see Appendix \ref{subsec:allSystemsENHI}).

\subsection{Statistics of Bitext for secondary set of experiments}
\label{subsec:stats_others}
Table \ref{tab:orig-secondary-corpora} gives the statistics of the original bitext that we obtained for the secondary set of experiments, to see the transferability of asymmetric BPE configurations. And to simulate low-resource settings, we sampled 0.1M sentence pairs per language using sentence-length binning, as done for English–Hindi; statistics are shown in Table~\ref{tab:sampled-secondary-corpora}.
\begin{table}[h!]
\centering
\resizebox{\columnwidth}{!}{%
\begin{tabular}{|l|r|r|r|}
\hline
\textbf{Language} & \textbf{\# Sentence Pairs} & \textbf{English Tokens} & \textbf{L Tokens} \\ \hline
Telugu     & 508,557     & 9,277,916     & 6,861,361 \\ \hline
Shona      & 9,463,612   & 98,089,812    & 76,046,554 \\ \hline
Norwegian  & 1,454,765   & 22,223,984    & 20,541,537 \\ \hline
Kyrgyz     & 21,603,490  & 251,345,836   & 168,333,543 \\ \hline
Hausa      & 4,452,045   & 57,987,583    & 64,016,592 \\ \hline
Inuktitut  & 733,624     & 15,751,147    & 7,991,818 \\ \hline
\end{tabular}
}
\caption{Original corpus statistics English - L Language for secondary language pair.}
\label{tab:orig-secondary-corpora}
\end{table}
\begin{table}[]
\centering
\resizebox{0.65\columnwidth}{!}{%
\begin{tabular}{|c|c|c|}
\hline
\textbf{Language} & \textbf{English Tokens} & \textbf{L Tokens} \\ \hline
Telugu     & 2,471,877     & 1,919,321 \\ \hline
Shona      & 1,228,485     & 965,502 \\ \hline
Norwegian  & 1,791,571     & 1,641,309 \\ \hline
Kyrgyz     & 1,385,891     & 936,543 \\ \hline
Hausa      & 1,531,132     & 1,679,785 \\ \hline
Inuktitut  & 2,148,188     & 1,089,834 \\ \hline
\end{tabular}
}
\caption{Token statistics after sampling 0.1 million training sentence pairs per language pair (English - L).}
\label{tab:sampled-secondary-corpora}
\end{table}
\subsection{Validation and Test Set Statistics}
\label{subsec:validAndTest}
\begin{table}[h!]
\centering
\resizebox{0.9\columnwidth}{!}{%
\begin{tabular}{|c|c|c|c|c|}
\hline
\textbf{Language} & \textbf{Split} & \textbf{\# Sentences} & \textbf{English Tokens} & \textbf{L Tokens} \\ \hline
Hindi       & validation & 997   & 23,586  & 27,325  \\ \cline{2-5}
            & test       & 1,012 & 24,722  & 28,534  \\ \hline
Telugu      & validation & 997   & 23,586  & 19,443  \\ \cline{2-5}
            & test       & 1,012 & 24,722  & 20,213  \\ \hline
Shona       & validation & 997   & 23,586  & 19,116  \\ \cline{2-5}
            & test       & 1,012 & 24,722  & 19,958  \\ \hline
Norwegian   & validation & 997   & 23,586  & 23,472  \\ \cline{2-5}
            & test       & 1,012 & 24,722  & 24,213  \\ \hline
Kyrgyz      & validation & 997   & 23,586  & 18,935  \\ \cline{2-5}
            & test       & 1,012 & 24,722  & 20,022  \\ \hline
Hausa       & validation & 997   & 23,586  & 27,031  \\ \cline{2-5}
            & test       & 1,012 & 24,722  & 28,018  \\ \hline
Inuktitut   & validation & 5,433 & 66,431  & 37,321  \\ \cline{2-5}
            & test       & 6,139 & 86,661  & 47,813  \\ \hline
\end{tabular}
}
\caption{Validation and test set statistics for all language pairs.}
\label{tab:valtest-merged}
\end{table}
As noted, for English–Inuktitut validation and test sets, we use \citet{joanis-etal-2020-nunavut}. For all other language pairs, the FLORES dataset was used. Table~\ref{tab:valtest-merged} shows token-level statistics for validation and test sets across all language pairs.



\end{document}